\documentclass{article}

\usepackage[english]{babel}

\usepackage[a4paper,top=2cm,bottom=2cm,left=3cm,right=3cm,marginparwidth=1.75cm]{geometry}
\everymath{\displaystyle}

\newcommand{\Op}{\mathcal{G}^\dagger} 

\newcommand{\IR}{\mathbb{R}} 

\usepackage{amsmath}
\usepackage{amssymb}
\usepackage{mathtools}
\usepackage{microtype}
\usepackage{bbm}
\usepackage{graphicx}
\usepackage[colorlinks=true, allcolors=blue]{hyperref}
\usepackage{xcolor}
\usepackage{booktabs}
\usepackage{multicol}
\usepackage{subcaption}
\usepackage{enumitem}
\usepackage{tikz}
\usepackage[maxbibnames=9]{biblatex}
\usetikzlibrary{positioning, arrows.meta, fit, shapes.geometric, calc}
\usepackage{todonotes}
\usepackage{authblk}
\usepackage{lineno}
\usepackage{subfiles} 
\newcommand{\learningRate}{\eta}

\newcommand{\schedulerGamma}{\gamma}
\newcommand{\weightDecay}{\omega}
\newcommand{\width}{d_v}
\newcommand{\hiddenLayer}{L}
\newcommand{\fourierModes}{k_{max}}
\newcommand{\activation}{\sigma}
\newcommand{\paddingPoints}{n_{pad}}

\title{Translation Invariance of Neural Operators for the FitzHugh–Nagumo Model}

\author[1,2]{Luca Pellegrini}
\affil[1]{Department of Mathematics, University of Pavia, Pavia, Italy}
\affil[2]{Euler Institute, Faculty of Informatics, Università della Svizzera italiana, Lugano, Switzerland}
\date{}
\begin{document}

\maketitle

\hrulefill
\begin{abstract}
         
Neural operators (NOs) are powerful deep learning frameworks designed to learn solution operators of partial differential equations. This study evaluates the ability of NOs' to capture the stiff spatio-temporal dynamics of the FitzHugh-Nagumo model. A key contribution of this study is the assessment of  the translation invariance using a novel training strategy. Models are trained using an applied current with varying spatial locations and intensities at a fixed time, while the test set presents a challenging out-of-distribution scenario where the current is translated in both time and space. This approach significantly reduces dataset generation costs. We benchmark seven NO architectures: Convolutional Neural Operators (CNOs), Deep Operator Networks (DeepONets), DeepONets with CNN encoders, Proper Orthogonal Decomposition DeepONets, Fourier Neural Operators (FNOs), Tucker Tensorized FNOs, and Local Neural Operators. We evaluated these models based on their accuracy, efficiency, and inference speed. These results demonstrate that CNOs generalize well to translated test dynamics, whereas other architectures do not generalize well. On the training set, all architectures achieve comparable accuracy, with FNOs achieving the highest precision. However, this higher accuracy comes at an elevated computational cost. Meanwhile, DeepONets and their variants exhibit superior training and inference efficiency. These findings highlight the capabilities and limitations of NOs in modeling complex ionic dynamics and provide a comprehensive benchmark for scenarios involving translated dynamics.

    \noindent
    \\[0.4cm]
\end{abstract}
\hrulefill

\section{Introduction}
The FitzHugh-Nagumo (FHN) model \cite{fitzhugh1961impulses} is widely used in computational electrophysiology \cite{cebrian2024six,keener2009mathematical2}, with significant applications in both computational neuroscience \cite{izhikevich2007dynamical} and computational cardiology \cite{franzone2014mathematical}. Originally derived as a simplification of the biophysically detailed Hodgkin-Huxley model \cite{hodgkin1952quantitative}, the FHN system accurately captures the essential dynamics of action potential propagation while significantly reducing computational cost. This balance between physical fidelity and efficiency is why the model remains so widely used today \cite{cebrian2024six}. Mathematically, the system is characterized by a parabolic reaction-diffusion partial differential equation (PDE) coupled with a stiff ordinary differential equation (ODE). In recent years, there has been significant interest in applying Scientific Machine Learning (SciML) techniques to computational electrophysiology \cite{centofanti2024learning,ghafourpour2025noble,hofler2025physics,pellegrini2025learning,salvador2024whole,shekarpaz2024splitting,ziarelli2026learning}, that aim to  exploit the non-linear approximation capabilities of Artificial Neural Networks (ANNs) to solve both forward and inverse problems. Among the many ANN-based methods, Neural Operators (NOs) \cite{kovachki2023neural} represent a paradigm shift. These architectures are designed to learn mappings between infinite-dimensional function spaces. While numerous NOs architectures have been proposed in recent years, a comprehensive comparison of every available variant is practically infeasible. In particular, we consider the following: Convolutional Neural Operators (CNOs) \cite{CNO23raonic}, Deep Operator Networks (DONs) \cite{DON21lu} (including variants such as DONs with CNN encoders (DONs-CNN) \cite{kovachki2023neural}  and Proper Orthogonal Decomposition DON (POD-DONs) \cite{lu2022comprehensive}), Fourier Neural Operators (FNOs) \cite{FNO20li} and its variants Tucker Tensorized Fourier Neural Operators (TFNOs) \cite{zhou2026tuckerfno}, and Localized Neural Operators (LocalNOs) \cite{liu2024neural}. The aim of this work is to provide a comprehensive and rigorous comparison by evaluating the accuracy in both train and test sets, as well as the associated computational costs (training and inference). Furthermore, we exploit a fundamental physical property of the FHN model, which we will refer to as translation invariance. Given a constant set of parameters, the solution with respect to an applied current remains invariant regardless of its translation in time. We investigate whether NOs can intrinsically capture this property. Consequently, we propose a novel training strategy in which the training dataset is generated with an applied current that varies in spatial position and intensity while remaining fixed in time, while the test set introduces a more challenging out-of-distribution scenario in which the applied current is translated in both time and space. This approach enables us to assess which architectures effectively learn the underlying invariance property, which could help to reduce the computational burden of generating large-scale datasets.

The paper is organized as follows: Section \ref{section:methodology} defines the mathematical formulation of the FHN model and the novel training strategy that exploits translation invariance. Furthermore, we provide a formal introduction to NOs and to the specific architectures considered (CNOs, DONs, DONs-CNN, POD-DONs, FNOs, TFNOs, and LocalNOs). Section \ref{sec:Numerical results} presents the numerical results obtained for each architecture. Section \ref{sec:Comparision} provides an in-depth comparison of the models based on accuracy, computational efficiency, inference speed, and cost functions that will be introduced that take into account accuracy, training time, and the number of trainable parameters. Finally, Section \ref{sec:conclusion} summarizes our findings, and outlines directions for future research.

\section{Mathematical models and methods}\label{section:methodology}
The FHN model is described by a coupled system of a parabolic reaction-diffusion PDE for the transmembrane potential $V$ and a stiff ODE for the gating variable $w$.

\begin{equation}
    \begin{cases}
        \chi C_m \frac{\partial V}{\partial t} - \text{div}(D(x) \nabla V) + \chi I_{ion}(V,w) = I_{app}, \quad & \text{in } \Omega_T \\
        \frac{\partial w}{\partial t} - R(V,w) = 0, \quad                                                       & \text{in } \Omega_T \\
        \textbf{n}^T D_m \nabla V = 0, \quad                                                                    & \text{on } \Gamma_T \\
        V(x,0) = V_0(x), \quad w(x,0) = w_0(x), \quad                                                           & \text{in } \Omega
    \end{cases}
    \label{eq:monodomain}
\end{equation}
where $\Omega_T = \Omega \times (0,T]$. A more detailed description of the model is provided in Appendix \ref{sec:FHN model}. Throughout this work, we focus on the operator that maps the applied current $I_{app}$ to the solution $(V,w)$:

\begin{alignat}{2}\label{eq:operator_ion}
    \Op :  \mathcal{A} & \rightarrow \mathcal{D} \\
    I_{app}            & \mapsto (V,w) \notag
\end{alignat}
where the applied current is defined as a piecewise constant function:
\begin{equation}
    I_{app}(i,x,t) =
    \begin{cases}
        i, \ \ (x,t) \in \Omega_{T,stim} \\
        0, \ \ (x,t) \notin \Omega_{T,stim}
    \end{cases}
    \label{eq:applied_current}
\end{equation}

\begin{figure}[!ht]
    \centering
    \includegraphics[width= 0.7\textwidth]{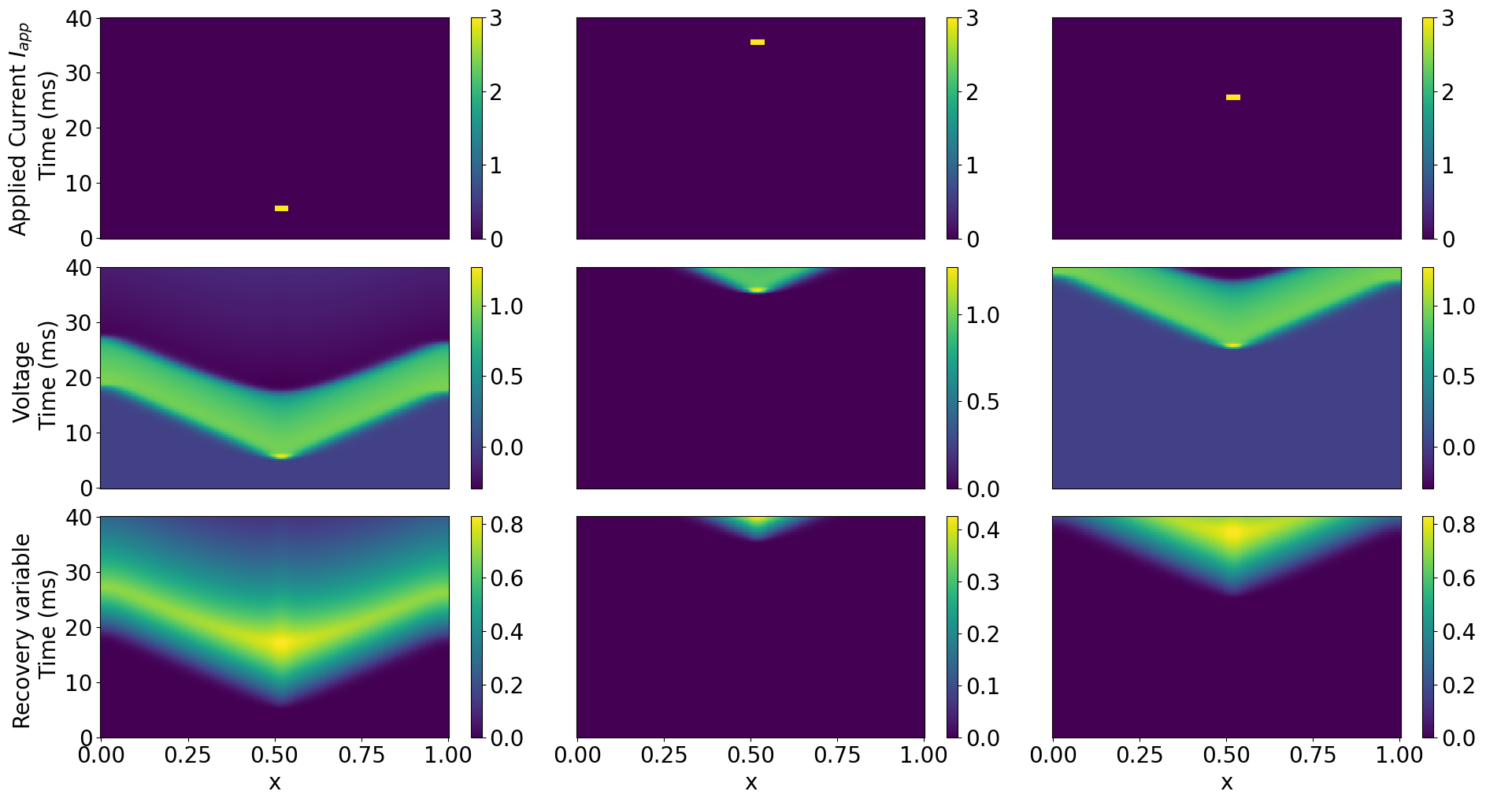}
    \caption{Example of translation invariance of the FHN model. Each column shows the evolution of the voltage V (second row) and the recovery variable w (third row) in response to an applied current $I_{app}$ \eqref{eq:applied_current} (first row), with an intensity of $i=3$, a duration of  1 ms, and spatial width 0.04 at 0.5. The stimulus $I_{app}$ is applied at three different times: 5 ms (first column), 35 ms (second column), and 25 ms (third column). These results were obtained using the Firedrake finite element library \cite{FiredrakeUserManual}.}
    \label{fig:fhn_invariance}
\end{figure}

Here, $\Omega_{T,stim} = \Omega_{stim} \times[T_{min},T_{max}]$, where $\Omega_{stim} \subset \Omega$ represents the region affected by the stimulus. As previously underlined, the FHN model exhibits a property that we refer to as translation invariance. This implies that, for a fixed set of parameters for the FHN model \eqref{eq:monodomain}, stimulus intensity $i$, and spatial position $\Omega_{stim}$, shifting the stimulus with respect to time results in an identical solution that has merely translated in time. This behaviour is illustrated in Figure \ref{fig:fhn_invariance}.

\begin{figure}[!ht]
    \centering
    \includegraphics[width= 0.7\textwidth]{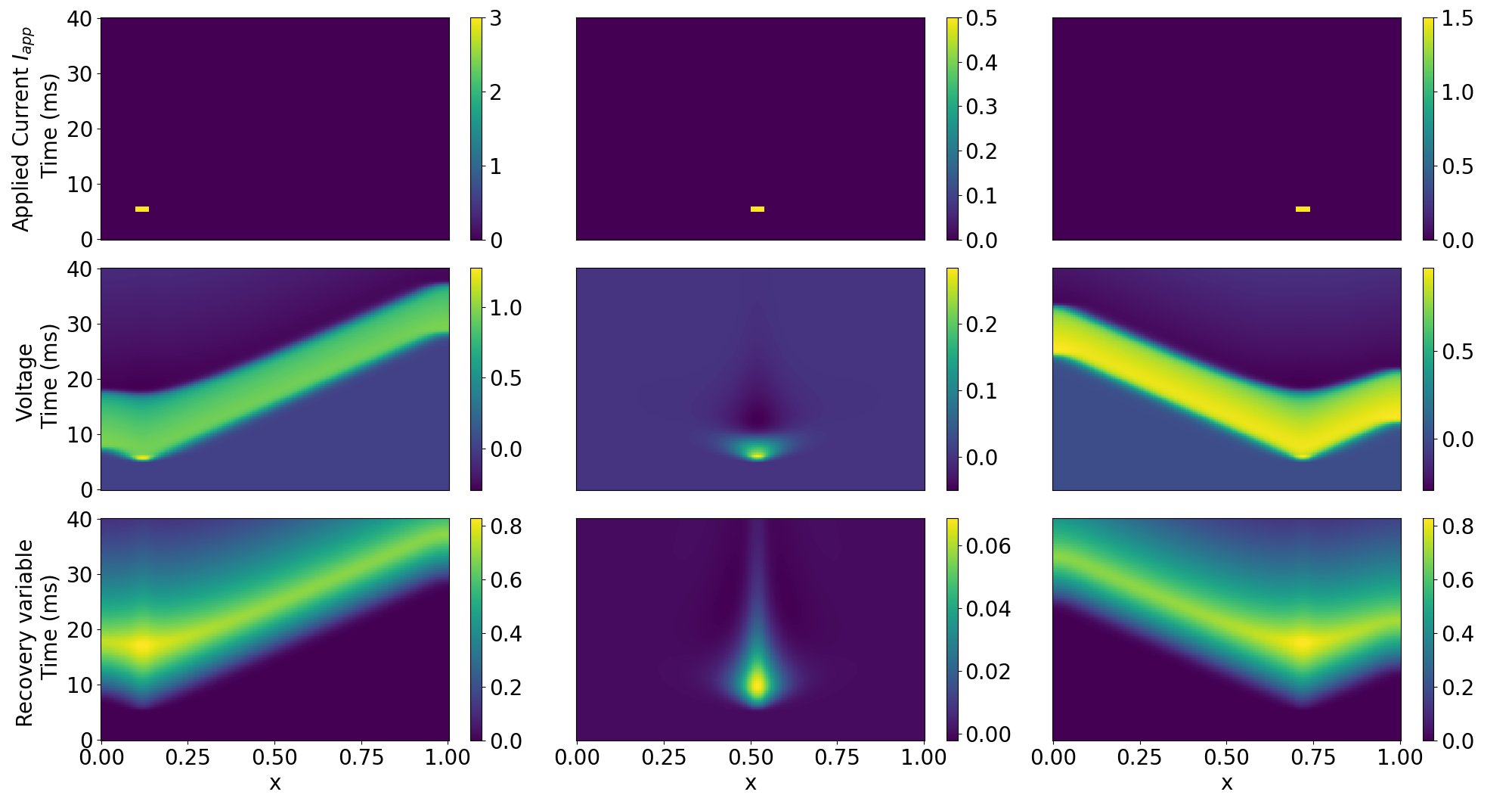}
    \caption{Examples from the training dataset for the FHN model. Each column illustrates the evolution of the voltage V (second row) and the recovery variable w (third row) in response to an applied current $I_{app}$ \eqref{eq:applied_current} (first row). In particular, the first column has an intensity $i=3$ at position 0.1, the second column has intensity of $i=0.5$ at position 0.5, and the third column has an intensity of $i= 1.5$ at position 0.7.  The stimulus begins at the same time for all three examples and has the same duration and spatial width: 5 ms, 1 ms, and 0.04, respectively. These results were obtained using the Firedrake finite element library \cite{FiredrakeUserManual}.}
    \label{fig:fhn_training}
\end{figure}

Our objective is to leverage this invariance by generating a training set in which the stimulus intensity $i$ and the spatial position $\Omega_{stim}$ are varied, while keeping the stimulus time ($T_{min}$ and $T_{max}$) fixed. This strategy is depicted in Figure \ref{fig:fhn_training}.

\begin{figure}[!ht]
    \centering
    \includegraphics[width= 0.7\textwidth]{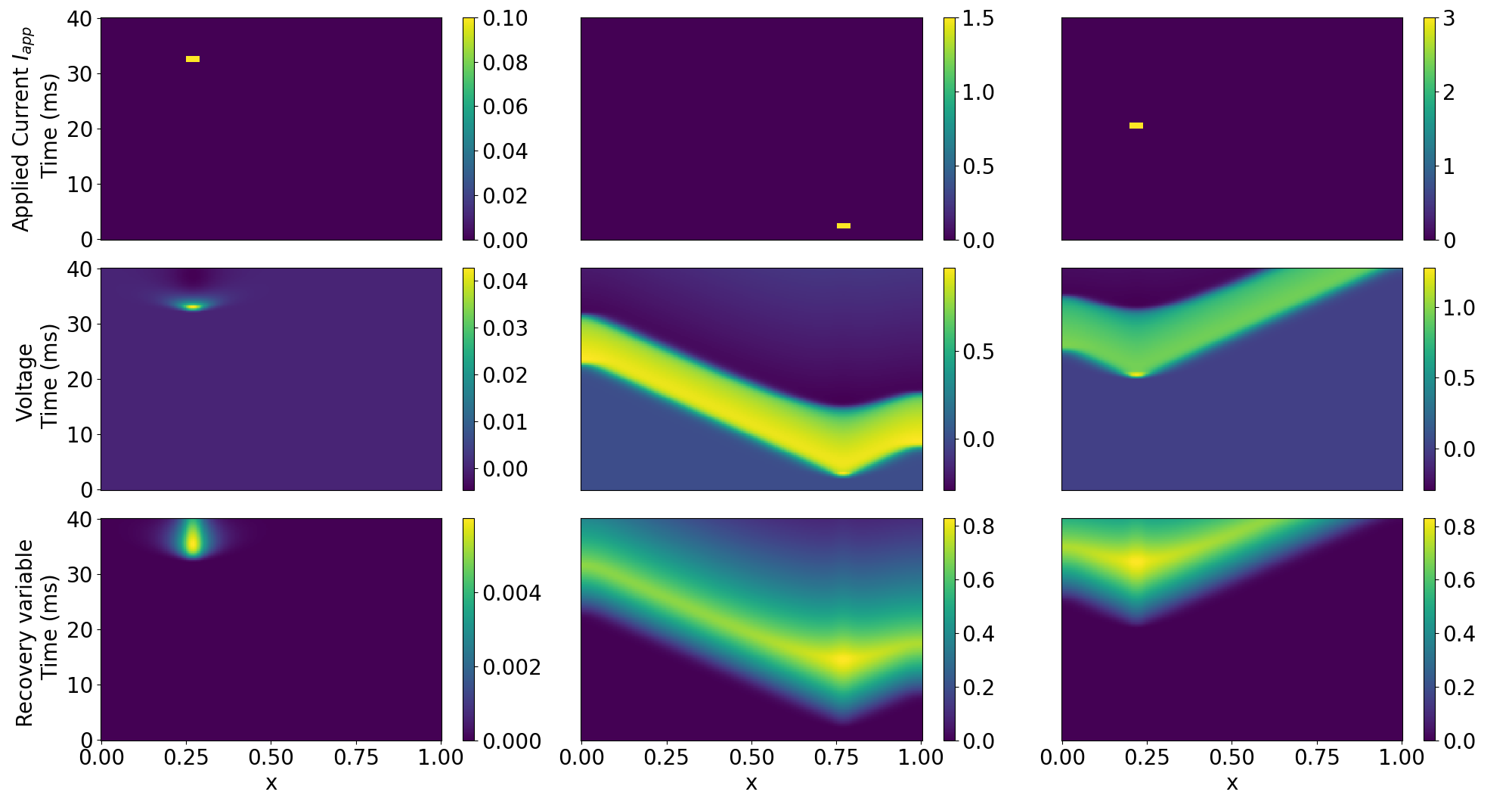}
    \caption{Examples from the test and validation datasets for the FHN model. Each column illustrates the evolution of the voltage V (second row) and the recovery variable w (third row) in response to an applied current $I_{app}$ \eqref{eq:applied_current} (first row). In particular, the first column has an intensity $i=0.1$ at $t=32$ ms and position $0.25$; the second column has an intensity $i=0.5$ at $t=2$ ms and position $0.5$; and the third column has an intensity $i=1.5$ at $t=20$ ms and position $0.7$. For all three cases, the stimulus duration and spatial width are held constant at 1 ms and 0.04, respectively. These results were obtained using the Firedrake finite element library \cite{FiredrakeUserManual}.}
    \label{fig:fhn_test}
\end{figure}

For the test and validation datasets, however, we vary the stimulus intensity $i$ and the spatial-temporal position $\Omega_{T,stim}$ (see Figure \ref{fig:fhn_test}).

This approach significantly reduces the computational cost of generating data by eliminating the need to sample the stimulus at different times. In this study, we evaluate the capacity of several NOs architectures to learn the translation invariance property. More details about the dataset generation can be found in the Appendix \ref{app_sec:dataset}.

\subsection{Neural Operators} \label{subsec: NO}
NOs have emerged as a promising framework for learning mappings between infinite-dimensional function spaces, making them particularly well-suited for solving PDEs \cite{neuraloperator21kov}. Unlike traditional neural networks, which operate on finite-dimensional spaces, NOs can learn continuous operators that map between function spaces. This section introduces NOs following the formalism presented in  \cite{li2,neuraloperator21kov}, for the FHN model \eqref{eq:monodomain}. In particular, we can rewrite our system Eq. \eqref{eq:monodomain} as follows:
\begin{equation*}
    \begin{cases}
        \mathcal{M} V\,(I_{app},V,w;x,t)= 0, \quad & (x,t)\in \Omega_T \\
        \mathcal{N} w\,(V,w;x,t) = 0,  \quad       & (x,t)\in \Omega_T \\
        \text{Boundary and Initial conditions.}
    \end{cases}
\end{equation*}
As previously mentioned, we are interested in the mapping defined by Eq. \eqref{eq:operator_ion}. Despite the fact that the applied current is explicitly present only in the first equation, it has an effect on both variables due to the coupled nature of the model.  Therefore, the solution operator is defined as $\Op  = \Phi^{-1}_{I_{app}}$ where $\Phi_{I_{app}}:=(\mathcal{M},\mathcal{N}) $. Furthermore, it is essential to emphasize that the two operators, denoted by $\mathcal{M}$ and $\mathcal{N}$, exhibit distinct scales and dynamics, which makes learning the dynamics of stiff ionic models, with NOs particularly challenging \cite{centofanti2024learning,pellegrini2025learning}. In general, NOs can be expressed as a composition of different operators:
\begin{equation}
    \mathcal{G}_{\theta} := \mathcal{Q} \circ \mathcal{L}_L \circ \cdots \circ \mathcal{L}_1 \circ \mathcal{P}:\ (I_{app},\theta) \mapsto \ (V,w)
    \label{eq:neural_operator_map}
\end{equation}
where the three main components are:
\begin{itemize}
    \item A lifting operator $\mathcal{P}$ which maps the input function (i.e. the applied current $I_{app}$) to a higher-dimensional space. This operator is typically implemented either as a shallow pointwise Multilayer Perceptron (MLP) or a single linear layer.
    \item A composition of integral operators $ \mathcal{L}_{\ell}$ processing the lifted representation through \( L \) layers in order to capture global dependencies and interactions in the input function through a sequence of transformations. Each $\mathcal{L}_\ell$ has the following structure:
          \begin{equation}\label{eq:integral_operator_fno}
              \mathcal{L}_{\ell}(v)(y) := \sigma\Big( W_{\ell} v(x)+ b_{\ell} + \mathcal{K}_{\ell}(\theta_{\ell}) v(y) \Big)
          \end{equation}
          where $v$ is the output of the previous layer, $W_{\ell}$ is a matrix, $b_{\ell}$ is a vector, $\theta_{\ell} \in \Theta_{\ell} \subset \Theta$ is a subset of the trainable parameters, and $\sigma$ is a non-linear activation function. Finally, $\mathcal{K}_{\ell}(\theta_{\ell})$ is a trainable linear operator.
    \item A projection operator $\mathcal{Q}$ which maps the internal representation back to the output space. This operator is also implemented as a pointwise MLP.
\end{itemize}
For a more thorough description of the mathematical properties of NOs, the reader is referred to \cite{bartolucci2023neural,lanthaler2023nonlocal}. In the following sections we will give a brief introduction to the NO architecture that will be taken into consideration throughout this work (i.e. CNOs, DONs, DONs-CNN, POD-DONs,  FNOs, TFNOs, and LocalNOs).

\subsubsection{Convolutional Neural Operators}
CNOs \cite{CNO23raonic} are a modification of Convolutional Neural Networks (CNNs) designed to enforce the structure-preserving continuous equivalence and ensure the architecture is a representation-equivalent neural operator \cite{neuraloperator21kov}, which are fundamental aspects of NOs. While a detailed derivation is beyond the scope of this work, we outline the essential components here, a complete technical definition is reported in \cite{CNO23raonic}. The mathematical foundation of CNOs is built upon the space of bandlimited functions, which are defined as follows:

\[\mathcal{B}_\omega(D,\IR^{d_v})\coloneqq \{f\in L^2(D,\IR^{d_v}: \text{supp}[\mathcal{F}(f)]\subset [-\omega,\omega]^2)\}.\]

Where, $\mathcal{F}$ is the Fourier transform and $\omega>0$ is the bandlimit frequency. The integral operator Eq. \refeq{eq:integral_operator_fno} in the case of CNO is defined as follows:

\[v_{\ell+1}=\mathcal{L}_{\ell}(v_{\ell})\coloneqq \mathcal{P}_\ell \circ \Sigma_\ell \circ \mathcal{K}_\ell(v_\ell).\]
Here, $\mathcal{P}_\ell$ is either the upsampling or downsampling operator, $\mathcal{K}_\ell$ is the convolution operator and $\Sigma_\ell$ is the activation operator. Following this, we will give a brief overview of the CNO's components, for more details refer to \cite{ghiotto2025hypernosautomatedparallellibrary,CNO23raonic,}.

\begin{figure}[!ht]
    \centering
    \begin{tikzpicture}[
            box/.style n args={2}{
                    draw,
                    align=center,
                    minimum height=#1,
                    minimum width=#2,
                    text width=#2,
                    inner sep=0pt,
                    fill=orange!30
                },
            greenbox/.style n args={2}{
                    draw,
                    align=center,
                    minimum height=#1,
                    minimum width=#2,
                    text width=#2,
                    inner sep=0pt,
                    fill=green!25
                },
            graybox/.style n args={2}{
                    draw,
                    align=center,
                    minimum height=#1,
                    minimum width=#2,
                    text width=#2,
                    inner sep=0pt,
                    fill=gray!50
                },
            conv/.style={-stealth, blue!70, thick},
            inv/.style={-stealth, violet, thick},
            connection/.style={-stealth, black, thick},
            pool/.style={-stealth, red!60, thick},
            upconv/.style={-stealth, green!60!black, thick},
            copy/.style={-stealth, gray!50, line width=0.1cm},
            copynoarrow/.style={gray!50, line width=0.1cm},
            label/.style={font=\scriptsize, align=center},
            legend/.style={font=\footnotesize, align=left}
        ]

        \def\boxwidth{0.1cm}
        \def\boxheight{3cm}
        \def\scale{0.5}

        \def\bufflabel{4}

        \def\spacing{0.4cm}
        \def\vspacing{0.7cm}

        \def\buffarrow{0.05cm}

        \node[box={\boxheight}{\boxwidth}] (l1) at (-\spacing,0) {};
        \node[label, rotate=90] at (-\boxwidth/2-1.5*\bufflabel-\spacing, 0) {$64^2\times 1$};

        \node[box={\boxheight}{\boxwidth}] (l2) at (\boxwidth+\spacing,0) {};

        \node[box={\boxheight}{2*\boxwidth}] (l3) at (2*\boxwidth+2*\spacing,0) {};
        \node[label,anchor=south] at (l3.north) {$64^2\times C$};

        \node[box={\boxheight*\scale}{2*\boxwidth}] (l4) at (2*\boxwidth+2*\spacing,-\boxheight/2 -\boxheight*\scale/2-\vspacing) {};
        \node[label, rotate=90] at ($(l4)+(-\boxwidth-1.5*\bufflabel, 0)$) {$32^2\times C$};

        \node[box={\boxheight*\scale}{2*\boxwidth}] (l5) at (4*\boxwidth+3*\spacing,-\boxheight/2 -\boxheight*\scale/2-\vspacing) {};

        \node[box={\boxheight*\scale}{4*\boxwidth}] (l6) at (7*\boxwidth+4*\spacing,-\boxheight/2 -\boxheight*\scale/2-\vspacing) {};
        \node[label,anchor=south] at (l6.north) {$32^2\times 2C$};

        \node[box={\boxheight*\scale^2}{4*\boxwidth}] (l7) at (7*\boxwidth+4*\spacing, -\boxheight/2 -\boxheight*\scale-\boxheight*\scale^2/2-2*\vspacing) {};
        \node[label] at ($(l7)+(-2*\boxwidth-5*\bufflabel, 0)$) {$16^2\times 2C$};

        \node[box={\boxheight*\scale^2}{4*\boxwidth}] (l8) at (11*\boxwidth+5*\spacing,-\boxheight/2 -\boxheight*\scale-\boxheight*\scale^2/2-2*\vspacing) {};

        \node[box={\boxheight*\scale^2}{8*\boxwidth}] (l9) at (17*\boxwidth+6*\spacing,-\boxheight/2 -\boxheight*\scale-\boxheight*\scale^2/2-2*\vspacing) {};
        \node[label,anchor=south] at (l9.north) {$16^2\times 4C$};

        \node[box={\boxheight*\scale^3}{8*\boxwidth}] (b1) at (17*\boxwidth+6*\spacing,-\boxheight/2 -\boxheight*\scale-\boxheight*\scale^2-\boxheight*\scale^3/2-3*\vspacing) {};
        \node[label] at ($(b1)+(-4*\boxwidth-5*\bufflabel, 0)$) {$8^2\times 4C$};

        \node[box={\boxheight*\scale^3}{8*\boxwidth}] (b2) at (25*\boxwidth+7*\spacing,-\boxheight/2 -\boxheight*\scale-\boxheight*\scale^2-\boxheight*\scale^3/2-3*\vspacing) {};



        \node[box={\boxheight*\scale^3}{8*\boxwidth}] (r2) at (38*\boxwidth+13*\spacing,-\boxheight/2 -\boxheight*\scale-\boxheight*\scale^2-\boxheight*\scale^3/2-3*\vspacing) {};

        \node[box={\boxheight*\scale^3}{8*\boxwidth}] (r3) at (46*\boxwidth+14*\spacing,-\boxheight/2 -\boxheight*\scale-\boxheight*\scale^2-\boxheight*\scale^3/2-3*\vspacing) {};

        \node[box={\boxheight*\scale^2}{8*\boxwidth}] (r4) at (46*\boxwidth+14*\spacing,-\boxheight/2 -\boxheight*\scale-\boxheight*\scale^2/2-2*\vspacing) {};
        \node[greenbox={\boxheight*\scale^2}{8*\boxwidth}] (r4-l) at (38*\boxwidth+14*\spacing,-\boxheight/2 -\boxheight*\scale-\boxheight*\scale^2/2-2*\vspacing) {};

        \node[box={\boxheight*\scale^2}{4*\boxwidth}] (r5) at (51*\boxwidth+15*\spacing,-\boxheight/2 -\boxheight*\scale-\boxheight*\scale^2/2-2*\vspacing) {};

        \node[box={\boxheight*\scale^2}{4*\boxwidth}] (r6) at (54*\boxwidth+16*\spacing,-\boxheight/2 -\boxheight*\scale-\boxheight*\scale^2/2-2*\vspacing) {};

        \node[box={\boxheight*\scale}{4*\boxwidth}] (r7) at (54*\boxwidth+16*\spacing,-\boxheight/2 -\boxheight*\scale/2-\vspacing) {};
        \node[greenbox={\boxheight*\scale}{4*\boxwidth}] (r7-l) at (50*\boxwidth+16*\spacing,-\boxheight/2 -\boxheight*\scale/2-\vspacing) {};

        \node[box={\boxheight*\scale}{2*\boxwidth}] (r8) at (56*\boxwidth+17*\spacing,-\boxheight/2 -\boxheight*\scale/2-\vspacing) {};

        \node[box={\boxheight*\scale}{2*\boxwidth}] (r9) at (58*\boxwidth+18*\spacing,-\boxheight/2 -\boxheight*\scale/2-\vspacing) {};

        \node[graybox={\boxheight}{\boxwidth}] (res1) at (26*\boxwidth+5*\spacing,0) {};
        \node[graybox={\boxheight}{\boxwidth}] (res2) at (27*\boxwidth+6*\spacing,0) {};
        \node[graybox={\boxheight}{\boxwidth}] (res3) at (28*\boxwidth+7*\spacing,0) {};
        \node[graybox={\boxheight}{\boxwidth}] (res4) at (29*\boxwidth+9*\spacing,0) {};
        \node[graybox={\boxheight}{\boxwidth}] (res5) at (30*\boxwidth+10*\spacing,0) {};
        \node[graybox={\boxheight}{\boxwidth}] (res6) at (31*\boxwidth+11*\spacing,0) {};
        \node[graybox={\boxheight}{\boxwidth}] (res7) at (32*\boxwidth+13*\spacing,0) {};
        \node[graybox={\boxheight}{\boxwidth}] (res8) at (33*\boxwidth+14*\spacing,0) {};
        \node[graybox={\boxheight}{\boxwidth}] (res9) at (34*\boxwidth+15*\spacing,0) {};
        \node[draw, line width = .7pt, rounded corners, inner sep=0.45cm, fit= (res1) (res2) (res3) (res4) (res5) (res6) (res7) (res8) (res9)] (internal) {};
        \node[label,anchor=south] at (internal.north) {$\textcolor{black}{N_{res}}$};

        \node[box={\boxheight}{2*\boxwidth}] (f1) at (58*\boxwidth+18*\spacing,0) {};
        \node[greenbox={\boxheight}{2*\boxwidth}] (f1-l) at (56*\boxwidth+18*\spacing,0) {};

        \node[box={\boxheight}{\boxwidth}] (f2) at (59*\boxwidth+19*\spacing,0) {};

        \node[box={\boxheight}{\boxwidth}] (f3) at (60*\boxwidth+20*\spacing,0) {};

        \node[box={\boxheight}{\boxwidth}] (f4) at (61*\boxwidth+22*\spacing,0) {};

        \draw[connection] (l1) -- (l2) node[midway, above] {$\mathcal{P}$};
        \draw[conv] (l2) -- (l3);
        \draw[pool] (l3) -- (l4);
        \draw[conv] (l4) -- (l5);
        \draw[conv] (l5) -- (l6);
        \draw[pool] (l6) -- (l7);
        \draw[conv] (l7) -- (l8);
        \draw[conv] (l8) -- (l9);
        \draw[pool] (l9) -- (b1);

        \draw[conv] (b1) -- (b2);

        \draw[conv] (r2) -- (r3);
        \draw[upconv] (r3) -- (r4);
        \draw[inv] (r4) -- (r5);
        \draw[conv] (r5) -- (r6);
        \draw[upconv] (r6) -- (r7);
        \draw[inv] (r7) -- (r8);
        \draw[conv] (r8) -- (r9);
        \draw[upconv] (r9) -- (f1);
        \draw[inv] (f1) -- (f2);
        \draw[conv] (f2) -- (f3);
        \draw[connection] (f3) -- (f4) node[midway, above] {$\mathcal{Q}$};

        \draw[copynoarrow] ($(l3) + (\boxwidth+0.2, 0)$) -- ($(res1) - (0.52, 0)$);
        \draw[copy] ($(res9) + (0.52, 0)$) -- ($(f1-l) - (\boxwidth +0.2 , 0)$);
        \draw[copy] ($(l6) + (2*\boxwidth+0.2, 0)$) -- ($(r7-l) - (2*\boxwidth+0.2, 0)$) node[label,midway, above] {$\textcolor{black}{N_{res}}$};
        \draw[copy] ($(l9) + (4*\boxwidth+0.2, 0)$) -- ($(r4-l) - (4*\boxwidth+0.2, 0)$) node[label,midway, above] {$\textcolor{black}{N_{res}}$};
        \draw[copy] ($(b2) + (4*\boxwidth+0.2, 0)$) -- ($(r2) - (4*\boxwidth+0.2, 0)$) node[label,midway, above] {$\textcolor{black}{N_{res,neck}}$};;

        \draw[connection] ($(res1) - (0.5, 0)$)-- (res1);
        \draw[conv] (res1) -- (res2);
        \draw[conv] (res2) -- (res3);
        \draw[connection] ($(res1) + (-0.25, 0)$) .. controls +(-0.5,2.4) and +(0.5,2.4) .. ($(res3) + (0.25, 0)$);
        \draw[connection] (res3) -- (res4);
        \draw[conv] (res4) -- (res5);
        \draw[conv] (res5) -- (res6);
        \draw[connection] ($(res4) + (-0.25, 0)$) .. controls +(-0.5,2.4) and +(0.5,2.4) .. ($(res6) + (0.25, 0)$);
        \draw[connection] (res6) -- (res7);
        \draw[conv] (res7) -- (res8);
        \draw[conv] (res8) -- (res9);
        \draw[connection] ($(res7) + (-0.25, 0)$) .. controls +(-0.5,2.4) and +(0.5,2.4) .. ($(res9) + (0.25, 0)$);
        \draw[connection] (res9) -- ($(res9) + (0.5, 0)$);

        \begin{scope}[shift={(58*\boxwidth+17*\spacing,-\boxheight/2 -\boxheight*\scale-\boxheight*\scale^2/2-2*\vspacing)}]
            \draw[conv] (0,0) -- (1,0) node[legend, right] {Convolutional} node[legend, black, yshift=10, xshift=-2] {Blocks legend:};
            \draw[copy] (0,-0.5) -- (1,-0.5) node[legend, right] {Residual};
            \draw[pool] (0,-1) -- (1,-1) node[legend, right] {Downsampling};
            \draw[upconv] (0,-1.5) -- (1,-1.5) node[legend, right] {Upsampling};
            \draw[inv] (0,-2) -- (1,-2) node[legend, right] {Invariant};
        \end{scope}
    \end{tikzpicture}
    \caption{Visual representation of a Convolutional Neural Operator with channel multiplier $C$ and initial resolution equal to $64$ with four layers.}
    \label{fig:cno}
\end{figure}
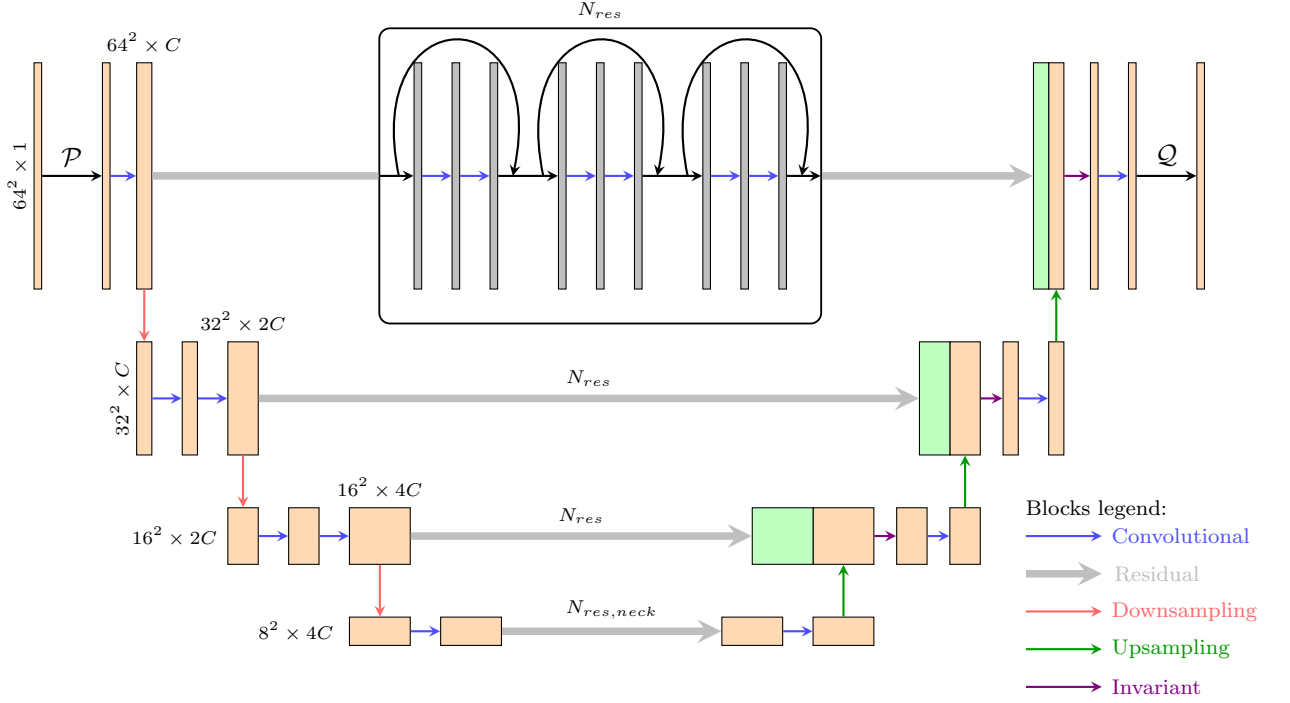

\begin{itemize}
    \item Upsampling $\mathcal{U}_{\omega,\bar{\omega}}$: This operator takes a function $f\in \mathcal{B}_\omega(D,\IR)$ and upsamples it to a higher bandlimit frequency space $\mathcal{B}_{\bar{\omega}}(D,\IR)$, where $\bar{\omega}>\omega$. Since $\mathcal{B}_{\omega}(D,\IR)\subset \mathcal{B}_{\bar{\omega}}(D,\IR)$ , the operation is simply the identity:
          \[\mathcal{U}_{\omega,\bar{\omega}} f(x) = f(x), \quad \forall x \in D.\]

    \item Downsampling $\mathcal{D}_{\omega,\underline{\omega}}$: This operator takes a function $f\in \mathcal{B}_\omega(D,\IR)$ and downsamples it to a lower bandlimit frequency space $\mathcal{B}_{\underline{\omega}}(D,\IR)$, where $\omega>\underline{\omega}$. It is defined as follows:
          \[\mathcal{D}_{\omega,\underline{\omega}} f(x) = \Big(\frac{\underline{\omega}}{\omega}\Big)^2(h_{\underline{\omega}}*f)(x).\]
          Where, $h_{\underline{\omega}}$ is the sinc filter defined as:
          \[h_{\omega}(x)=\prod_{i=1}^{d}\frac{sin(2\pi \omega x_i)}{2 \pi \omega x_i}, \quad \forall x \in \IR^d.\]

    \item Activation operator $\Sigma_{\omega}$: This operator is designed such that it maps functions from a bandlimited space $\mathcal{B}_{\omega}(D,\IR)$ back to the same space (i.e., if $f\in\mathcal{B}_{\omega}(D,\IR) \text{ then }\Sigma_{\omega}(f)\in \mathcal{B}_{\omega}(D,\IR$)). It is defined as:
              \[\Sigma_\omega f(x) = (\mathcal{D}_{\bar{\omega},\omega} \circ \sigma \circ \mathcal{U}_{\omega,\bar{\omega}} f )(x), \quad \forall x \in D\]
              Where, $\sigma$ is an activation function, and $\bar{\omega}$ is chosen large enough such that $\sigma(\mathcal{B}_{\omega}(D,\IR))\subset \mathcal{B}_{\bar{\omega}}(D,\IR)$

    \item Convolution Operator $\mathcal{K}_\ell$: With a slight abuse of notation, we denote $\mathcal{K}_\ell$ as $\mathcal{K}_{\omega,\theta}$, where $\omega$ is the bandwidth and $\theta$ are the learnable parameters. We assume there is a function $\kappa_{\omega,\theta}$ called the kernel function, such that the operator can be rewritten as the following convolution:
          \[\mathcal{K}_{\omega,\theta}f(x)=(f*\kappa_{\omega,\theta})(x) = \int_{D} \kappa_{\omega,\theta}(y)f(x-y)dy, \quad f\in\mathcal{B}_\omega (D,\IR^{d_v}), \quad \forall x \in  D.\]
          The integral can then be approximated as:
          \[\mathcal{K}_{\omega,\theta}f(x) \approx \sum_{i,j= 1}^{k} \kappa_{\omega,\theta}(y_{ij})f(x-y_{ij}) \quad \forall x\in D.\]
          Where in this context, $k$ is the kernel size of the convolution and $d_v$ is the channel dimension.

\end{itemize}
If we are dealing with a vector-valued function $f\in \mathcal{B}_\omega(D,\IR^{d_v})$, the operators are applied component-wise. The aforementioned operators are then assembled into a U-Net architecture \cite{ronneberger2015u} featuring skip connection, the two additional operators described below.
\begin{itemize}
    \item Skip Connection (ResNet Blocks) $\mathcal{R}_{\omega,\theta}$: These blocks connect the encoder and decoder layers at the same spatial resolution. They are useful for transferring high-frequency information to the output before it is filtered by the downsampling operation. The block is defined as:
          \[\mathcal{R}_{\omega,\theta}(v)= v + \mathcal{K}_{\omega,\theta}\circ \Sigma_\omega \circ \mathcal{K}_{\omega,\theta}(v), \quad \forall v \in \mathcal{B}_\omega(D,\IR^{d_v}).\]
          The output of the residual block is concatenated with the input of the corresponding decoder block.

    \item Invariant Blocks  $\mathcal{I}_{\omega,\theta}$: At every layer of the U-Net, after the concatenation of the residual block output with the decoder input, we apply an invariant block, defined as:
          \[\mathcal{I}_{\omega,\theta} = \Sigma_\omega \circ \mathcal{K}_{\omega,\theta}(v), \quad \forall v \in \mathcal{B}_\omega (D,\IR^{d_v})\]
\end{itemize}
A visual representation of the CNO architecture is shown in Fig \ref{fig:cno}.
\subsubsection{Deep Operator Networks} \label{subsec:DONs}

DONs \cite{DON21lu} were among the first architectures proposed to learn the infinite-dimensional mappings arising from PDEs. However, it is important to observe that their vanilla formulation does not strictly correspond to the iterative structure defined by Eq. \eqref{eq:neural_operator_map}. However, as demonstrated in the recent study \cite{kovachki2023neural}, under certain assumptions,  DONs can be mathematically defined as NOs. In terms of architecture, a DON is defined as two sub-networks: the branch and the trunk. The branch takes as input the $I_{app}$ and outputs a set of coefficients, $[b_1,...,b_p]$, where $p$ denotes the number of basis functions. The trunk takes the spatio-temporal coordinates, $(x,t)$ as input and outputs the basis functions [$T_1,...T_p$]. The final output is obtained by merging these two components via a dot product.

\[\mathcal{G}_{\theta}(I_{app})(x,t) = \sum_{i=1}^{p} b_i(I_{app}) T_i(x,t), \quad \forall (x,t) \in \Omega_T .\]

Here, $\theta$ represents the trainable parameters of both the branch and trunk networks. To handle the coupled system of PDEs, one could potentially employ two separate DONs architectures and optimize them through a joint loss function. In this work, however, we employ a single DONs and split the output into two components. In particular, for a total of $p$ basis, we employ $p/2$ basis functions and their corresponding coefficients for the potential $V$, and the remaining $p/2$ for the gating variable $w$. This can be written as:

\[\mathcal{G}_{\theta}(I_{app})(x,t) = \underbrace{\sum_{i=1}^{p/2} b_i(I_{app}) T_i(x,t)}_{V_{\theta}} + \underbrace{\sum_{i=p/2+1}^{p} b_i(I_{app}) T_i(x,t)}_{w_\theta}, \quad \forall (x,t) \in \Omega_T .\]

\begin{figure}[!ht]
    \centering
    \begin{tikzpicture}[
            box_input/.style={draw, rounded corners, align=center, minimum height=0.8cm, minimum width=1cm, fill=orange!31},
            box/.style n args={2}{
                    draw,
                    align=center,
                    minimum height=#1,
                    minimum width=#2,
                    text width=#2,
                    inner sep=0pt,
                    fill=red!30
                },
            greenbox/.style n args={2}{
                    draw,
                    align=center,
                    minimum height=#1,
                    minimum width=#2,
                    text width=#2,
                    inner sep=0pt,
                    fill=green!25
                },
            conv/.style={-stealth, blue!70, thick},
            inv/.style={-stealth, violet, thick},
            connection/.style={-stealth, black, thick},
            pool/.style={-stealth, red!60, thick},
            upconv/.style={-stealth, green!60!black, thick},
            copy/.style={-stealth, gray!50, line width=0.1cm},
            copynoarrow/.style={gray!50, line width=0.1cm},
            label/.style={font=\scriptsize, align=center},
            legend/.style={font=\footnotesize, align=left},
            nn_node/.style={
                    circle, draw, fill=violet!25,
                    minimum size=4mm, inner sep=0pt
                },
            fnn_node/.style={
                    circle, draw , fill=blue!40,
                    minimum size=4mm, inner sep=0pt
                },
            fnn_connect/.style={
                    gray, thin
                }
        ]

        \def\boxwidth{0.3cm}
        \def\boxheight{2.5cm}
        \def\scale{0.5}

        \def\bufflabel{4}

        \def\spacing{0.7cm}
        \def\vspacing{0.7cm}

        \def\buffarrow{0.05cm}

        \def\xflat{5.5*\boxwidth+5.5*\spacing}
        \def\xhone{7*\boxwidth+7*\spacing}
        \def\xhtwo{8*\boxwidth+8*\spacing}
        \def\xhthree{9*\boxwidth+9*\spacing}
        \def\xout{10.5*\boxwidth+10.5*\spacing}


        \node[fnn_node] (h1-1) at (\xhone, 1.32) {};
        \node[fnn_node] (h1-2) at (\xhone, 0.58) {};
        \node[fnn_node] (h1-3) at (\xhone, -0.68) {};
        \node (flat-dots) at (\xhone, 0.105) {\(\vdots\)};

        \node[fnn_node] (h2-1) at (\xhtwo, 1.32) {};
        \node[fnn_node] (h2-2) at (\xhtwo, 0.58) {};
        \node[fnn_node] (h2-3) at (\xhtwo, -0.68) {};
        \node (flat-dots) at (\xhtwo, 0.105) {\(\vdots\)};

        \node[fnn_node] (h3-1) at (\xhthree, 1.32) {};
        \node[fnn_node] (h3-2) at (\xhthree, 0.58) {};
        \node[fnn_node] (h3-3) at (\xhthree, -0.68) {};
        \node at ($(h2-1)!0.5!(h3-1) - (0,1.32) $) {\(\cdots\)};





        \draw[fnn_connect] (h1-1) -- (h2-1);
        \draw[fnn_connect] (h1-1) -- (h2-2);
        \draw[fnn_connect] (h1-1) -- (h2-3);

        \draw[fnn_connect] (h1-2) -- (h2-1);
        \draw[fnn_connect] (h1-2) -- (h2-2);
        \draw[fnn_connect] (h1-2) -- (h2-3);

        \draw[fnn_connect] (h1-3) -- (h2-1);
        \draw[fnn_connect] (h1-3) -- (h2-2);
        \draw[fnn_connect] (h1-3) -- (h2-3);


        \node[draw, line width = .7pt, rounded corners, inner sep=0.45cm,
            fit= (h1-1)   (h3-3) ] (fnn) {};
        \node[label,anchor=south] at (fnn.north) {\textcolor{black}{\textit{FNN}}};

        \node[box_input, label=above:{\textit{Output}}] (output) at ($(fnn.east)+ (1.2cm,0)$) {Branch \\ Output};

        \draw[connection] (fnn) -- (output);

        \node[label,anchor=south] at ($(output.north) + (0, 0.2cm)$) {\textit{}};


        \def\uniformbranchspacing{1cm} 


        \node[box_input, label=above:{\textit{Input}}] (input) at ($(fnn.west)- (1.2cm,0)$) {$I_{app}(x,t)$};

        \node[label,anchor=south] at (input.north) {\textit{Input}};

        \draw[connection] (input) -- (fnn);


        \node[fnn_node] (th1-1) at (\xhone, 1.32-4) {};
        \node[fnn_node] (th1-2) at (\xhone, 0.58-4) {};
        \node[fnn_node] (th1-3) at (\xhone, -0.68-4) {};
        \node (flat-dots) at (\xhone, 0.105-4) {\(\vdots\)};

        \node[fnn_node] (th2-1) at (\xhtwo, 1.32-4) {};
        \node[fnn_node] (th2-2) at (\xhtwo, 0.58-4) {};
        \node[fnn_node] (th2-3) at (\xhtwo, -0.68-4) {};
        \node (flat-dots) at (\xhtwo, 0.105-4) {\(\vdots\)};

        \node[fnn_node] (th3-1) at (\xhthree, 1.32-4) {};
        \node[fnn_node] (th3-2) at (\xhthree, 0.58-4) {};
        \node[fnn_node] (th3-3) at (\xhthree, -0.68-4) {};
        \node at ($(th2-1)!0.5!(th3-1) - (0,1.32) $) {\(\cdots\)};

        \node (flat-dots) at (\xhthree, 0.105-4) {\(\vdots\)};

        \node[draw, line width = .7pt, rounded corners, inner sep=0.45cm,
            fit= (th1-1)  (th3-3)  ] (fnn_t) {};
        \node[label,anchor=south] at (fnn_t.north) {\textcolor{black}{\textit{FNN}}};
        \node[box_input, label=above:{\textit{Input}}] (input_t) at ($(fnn_t.west) +(-1.2,0)  $) {$(x,t)$};
        \node[label,anchor=south] at (input_t.north) {\textit{Input}};
        \node[box_input, label=above:{\textit{Output}}] (output_t) at ($(fnn_t.east)+ (1.2cm,0)$) {Trunk \\ Output};

        \draw[connection] (input_t) -- (fnn_t);
        \draw[fnn_connect] (th1-1) -- (th2-1);
        \draw[fnn_connect] (th1-1) -- (th2-2);
        \draw[fnn_connect] (th1-1) -- (th2-3);

        \draw[fnn_connect] (th1-2) -- (th2-1);
        \draw[fnn_connect] (th1-2) -- (th2-2);
        \draw[fnn_connect] (th1-2) -- (th2-3);

        \draw[fnn_connect] (th1-3) -- (th2-1);
        \draw[fnn_connect] (th1-3) -- (th2-2);
        \draw[fnn_connect] (th1-3) -- (th2-3);

        \draw[connection] (fnn_t) -- (output_t);

        \node[circle, draw,minimum size=4mm, inner sep=0pt] (output_final) at ($(output.south)!0.5!(output_t.north) $) {$\times$};
        \draw[fnn_connect] (output.south) -- (output_final);
        \draw[fnn_connect] (output_t.north) -- (output_final);
        \node[box_input] (outputs) at ($(fnn.south)!0.5!(fnn_t.north) + (5cm,0)$) {$(V,w)(x,t)$};
        \node[label,anchor=south] at (outputs.north) {\textit{Outputs}};
        \draw[connection] (output_final) -- (outputs);

    \end{tikzpicture}
    \caption{Visual representation of a DON architecture.}
    \label{fig:don}

\end{figure}
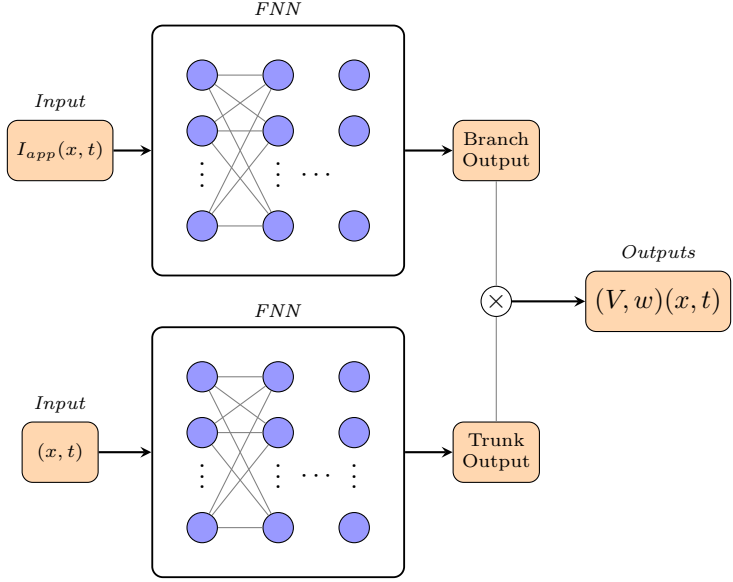

We adopted this single-architecture approach based on the idea that a unified network should capture the ionic model's intrinsic cross-dependencies and coupled dynamics. A visual representation of this architecture is shown in Fig. \ref{fig:don}.

Moreover, we will consider the following modification of the DON:

\begin{itemize}
    \item DONs with CNN encoders (DONs-CNN): In this variant, the input of the branch $I_{app}$ is not given directly to a fully connected ANN, but Instead, but instead first passes through a series of convolutional layers.
    \item Proper Orthogonal Decomposition DONs (POD-DONs) \cite{lu2022comprehensive}: This architecture combines the operator learning capabilities of DONs with the dimensionality reduction technique POD. Rather than using an ANN for the trunk network, the output is a linear combination of POD modes extracted from the training data.
\end{itemize}

\subsubsection{Fourier Neural Operators}\label{subsec:FNO}
FNOs \cite{neuraloperator21kov, FNO20li} are a class of NOs that employs the Fourier transform to efficiently parameterize the integral kernel operator $\mathcal{K}_{\ell}(\theta_{\ell})$ Eq. \eqref{eq:integral_operator_fno}. By operating in the frequency domain, FNOs effectively capture global dependencies within input functions. In the case of FNOs, the integral kernel operator is defined as follows:
\begin{equation}\label{eq:kernel}
    \mathcal{K}_{\ell,\theta_{\ell}} v(y)= \int_{\mathbb{T}^d} \kappa_{\ell,\theta_{\ell}}(y,s) v(s) \ ds = \int_{\mathbb{T}^d} \kappa_{\ell,\theta_{\ell}}(y-s) v(s) \ ds= (\kappa_{\ell, \theta_{\ell}} * v)(y)
\end{equation}
where $\mathbb{T}^d$ is the $d$-dimensional torus, and $\kappa_{\ell,\theta_{\ell}}$ is a kernel function, which depends on the learnable parameters $\theta_{\ell}$.
We can apply the convolution theorem for the Fourier transform in order to rewrite the convolution in \eqref{eq:kernel} in the Fourier domain:
\begin{equation*}
    (\kappa_{\ell, \theta_{\ell}} * v)(y) =  \mathcal{F}^{-1}\left( \mathcal{F}( \kappa_{\ell,\theta_{\ell}}) (k) \cdot \mathcal{F}(v)(k) \right)(y).
\end{equation*}
The key step in FNOs is the parameterization of $\mathcal{F}( \kappa_{\ell, \theta_{\ell}} )(k)$ using a complex-valued matrix of learnable parameters $R_{\theta_{\ell}}(k) \in \mathbb{C}^{d_v \times d_v}$ for each frequency mode $k \in \mathbb{Z}^d$, obtaining:
\begin{equation*}
    (\mathcal{K}_{\ell,\theta_{\ell}}v)(y)= \mathcal{F}^{-1}\left( R_{\theta_{\ell}}(k) \cdot \mathcal{F}(v)(k) \right)(y).
\end{equation*}
Since we consider real-valued functions, the parameterized matrix $R_{\theta_{\ell}}(k)$ must be Hermitian, i.e., $R_{\theta_{\ell}}(-k) = \overline{R_{\theta_{\ell}}(k)}$ for all $k \in \mathbb{Z}^d$ and $\ell=1,\ldots,L$. Furthermore, in order to apply the Fast Fourier Transform (FFT) for the numerical implementation of the Fourier transform, the mesh considered must be structured and uniform.
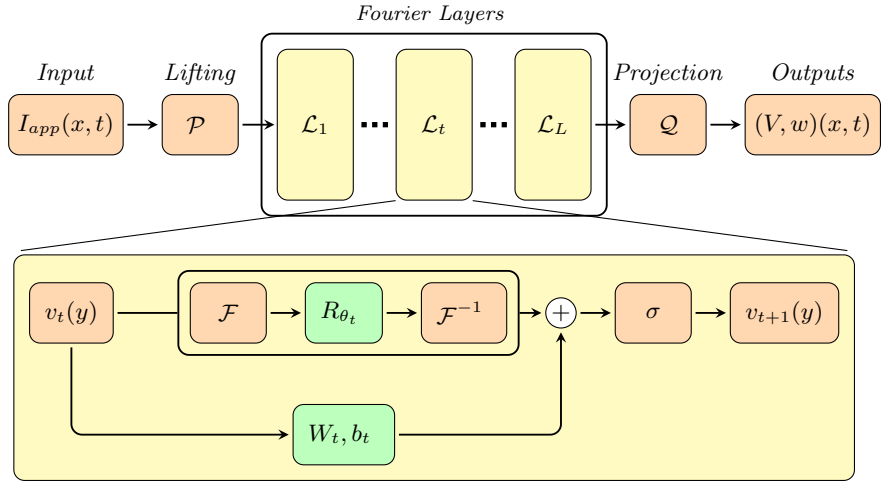
\begin{figure}[!ht]
    \centering
    \begin{tikzpicture}
        \tikzset{
            box/.style={draw, rounded corners, align=center, minimum height=0.8cm, minimum width=1cm, fill=orange!31},
            bigbox/.style={draw, rounded corners, align=center, minimum height=2cm, minimum width=1cm, fill=yellow!31},
            node_sum/.style={draw, circle, fill=white, inner sep=0pt, minimum size=4mm},
            every node/.style={font=\small}
        }
        \node[box, label=above:{\textit{Input}}] (input) {$I_{app}(x,t)$};
        \node[box, right=0.5cm of input, label=above:{\textit{Lifting}}] (lifting) { $\mathcal{P}$ };
        \node[bigbox, right=0.5cm of lifting] (fourier1) {$\mathcal{L}_{1}$};
        \node[bigbox, right=0.55cm of fourier1] (fourier2) {$\mathcal{L}_{t}$};
        \node[bigbox, right=0.55cm of fourier2] (fourier3) {$\mathcal{L}_{L}$};
        \node[box, right=0.5cm of fourier3, label=above:{\textit{Projection}}] (projection) {$\mathcal{Q}$};
        \node[box, right=0.5cm of projection, label=above:{\textit{Outputs}}] (output) {$(V,w)(x,t)$};

        \draw[-stealth, line width = .7pt] ($(input.east)+(0.05, 0)$) -- ($(lifting.west)-(0.05,0)$);
        \draw[-stealth, line width = .7pt] ($(lifting.east)+(0.05, 0)$) -- ($(fourier1.west)-(0.03,0)$);
        \draw[dotted, line width = 2pt] ($(fourier1.east)+(0.1, 0)$) -- ($(fourier2.west)-(0.1,0)$);
        \draw[dotted, line width = 2pt] ($(fourier2.east)+(0.1, 0)$) -- ($(fourier3.west)-(0.1,0)$);
        \draw[-stealth, line width = .7pt] ($(fourier3.east)+(0.05, 0)$) -- ($(projection.west)-(0.05,0)$);
        \draw[-stealth, line width = .7pt] ($(projection.east)+(0.05, 0)$) -- ($(output.west)-(0.05,0)$);

        \node[align=center, above=0.2cm of fourier2, font=\footnotesize] { \textit{Fourier Layers} };

        \node[draw, line width = .7pt, rounded corners, inner sep=0.2cm, fit= (fourier1) (fourier2) (fourier3)] (internal) {};

        \node[draw, below=0.5cm of internal, rounded corners, inner sep=0.2cm, fill=yellow!30] (internal) {
            \begin{tikzpicture}[every node/.style={font=\small}]
                \node[box] (vt) {$v_t(y)$};
                \node[box, right=1cm of vt] (transform) {$\mathcal{F}$};
                \node[box, right=0.5cm of transform, fill=green!26] (nonlinear) { $ R_{\theta_t} $ };
                \node[box, right=0.5cm of nonlinear] (invtransform) { $ \mathcal{F}^{-1} $ };
                \node[box, below=0.8cm of nonlinear, fill=green!26] (linear) { $ W_t, b_t $ };

                \node[draw, line width = .7pt, rounded corners, inner sep=0.15cm, fit= (transform) (nonlinear) (invtransform) ] (diagonalscaling) {};

                \node[node_sum, right=0.5cm of invtransform] (node_sum) {$\mathbf{+}$};
                \node[box, right=0.5cm of node_sum] (activation) {$\sigma$};
                \node[box, right=0.5cm of activation] (vtplusone) {$v_{t+1}(y)$};

                \draw[line width = .7pt] ($(vt.east)+(0.05, 0)$) -- (diagonalscaling);
                \draw[-stealth, line width = .7pt] ($(transform.east)+(0.05, 0)$) -- ($(nonlinear.west)-(0.05, 0)$);
                \draw[-stealth, line width = .7pt] ($(nonlinear.east)+(0.05, 0)$) -- ($(invtransform.west)-(0.05, 0)$);
                \draw[-stealth, line width = .7pt] ($(vt.south)-(0, 0.05)$) |- ($(linear.west)-(0.05, 0)$);
                \draw[-stealth, line width = .7pt] (diagonalscaling) -- ($(node_sum.west)-(0.05, 0)$);
                \draw[-stealth, line width = .7pt] ($(linear.east)+(0.05, 0)$) -| ($(node_sum.south)-(0, 0.05)$);
                \draw[-stealth, line width = .7pt] ($(node_sum.east)+(0.05, 0)$) -- ($(activation.west)-(0.05, 0)$);
                \draw[-stealth, line width = .7pt] ($(activation.east)+(0.05, 0)$) -- ($(vtplusone.west)-(0.05, 0)$);

            \end{tikzpicture}
        };
        \draw[] ($(internal.north west)+(0.1, 0.05)$) -- (fourier2.south west);
        \draw[] ($(internal.north east)+(-0.1, 0.05)$) -- (fourier2.south east);
    \end{tikzpicture}
    \label{fig:fno}
    \caption{Visual representation of a FNO.}
\end{figure}

In our implementation, we have considered two versions of \eqref{eq:integral_operator_fno}:
\begin{align}
    v_{\ell+1}^{Classic} & = \mathcal{L}_{\ell}^{Classic}(v_{\ell}) := \sigma\Big( W_{\ell} v_{\ell}+ b_{\ell} + \mathcal{K}_{\ell,\theta_{\ell}} v_{\ell} \Big), \label{eq:classic} \\
    v_{\ell+1}^{MLP}     & = \mathcal{L}_{\ell}^{MLP}(v_{\ell}) := \sigma\Big( W_{\ell} v_{\ell}+ b_{\ell} + MLP(\mathcal{K}_{\ell,\theta_{\ell}} v_{\ell} )\Big) \label{eq:mlp},
\end{align}
where $v_{\ell+1}^{\text{Classic}}$ has the exact structure described in \eqref{eq:integral_operator_fno}, while
$v_{\ell+1}^{\text{MLP}}$ generalizes the classic architecture by incorporating a pointwise MLP applied to the output of the integral kernel operator. Moreover, we will consider two modifications of the FNOs:
\begin{itemize}

    \item Local Neural Operators (LocalNOs) \cite{liu2024neural}: It employs localized integral kernels to capture high-frequency local features and discontinuities.

    \item Tucker Tensorized FNOs (TFNOs) \cite{zhou2026tuckerfno}: It combines Tucker tensor decomposition with FNOs by applying tensor decomposition to the Fourier weights, significantly reducing the model memory requirement and computational cost in high-dimensional spaces.
\end{itemize}

\section{Numerical Results}\label{sec:Numerical results}

In this section, we present the results of the numerical tests performed. We employed HyperNOs \cite{ghiotto2025hypernosautomatedparallellibrary}, a PyTorch library, which implements automatic hyperparameter tuning of NOs  based on Ray \cite{liaw2018tune}. All tests were performed on a Linux cluster. Two nodes, each with two NVIDIA A16 GPUs, were used for the automatic hyperparameter tuning. For single training, we used one node with an NVIDIA A16 GPU. For inference time, we used a laptop with an NVIDIA GeForce RTX 4070 GPU. More details about automatic hyperparameter tuning are reported in Appendix \ref{app_sec:ray}. In the following sections, we will provide a detailed description of the results obtained by each architecture. Each architecture has been trained using AdamW with weight decay regularization $\weightDecay$ for 1000 epochs. We also use a learning rate scheduler that decreases the learning rate $\learningRate$ by a factor of $\schedulerGamma$ every ten epochs. The dataset has 2000 examples for training and 500 for the test, more details are in the Appendix \ref{app_sec:dataset}.

\subsection{Convolutional Neural Operators results}\label{subsec:CNO_results}

\begin{figure}[!ht]
    \begin{subfigure}[t]{0.49\textwidth}
        \centering
        \includegraphics[width=0.9\textwidth]{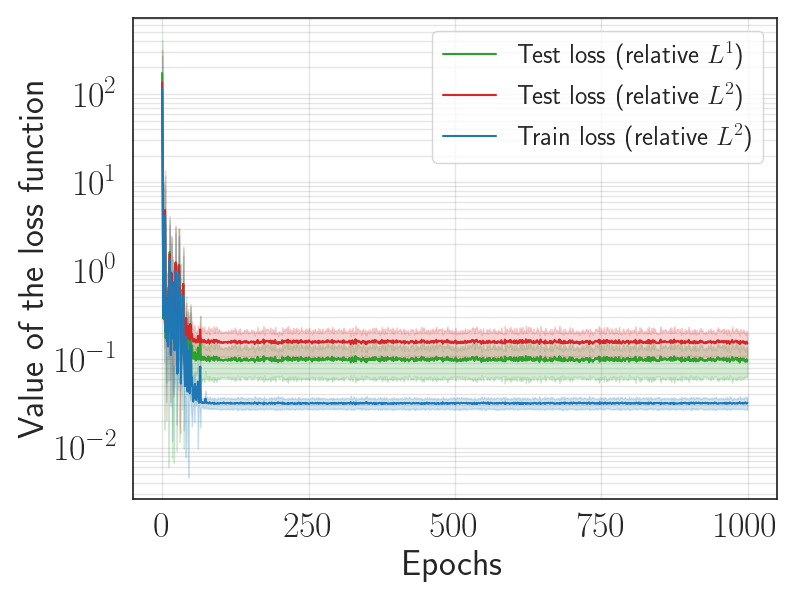}
        \caption{Loss values for the CNO.}
        \label{fig:loss_CNO}
    \end{subfigure}%
    \begin{subfigure}[t]{0.49\textwidth}
        \centering
        \includegraphics[width=0.9\textwidth]{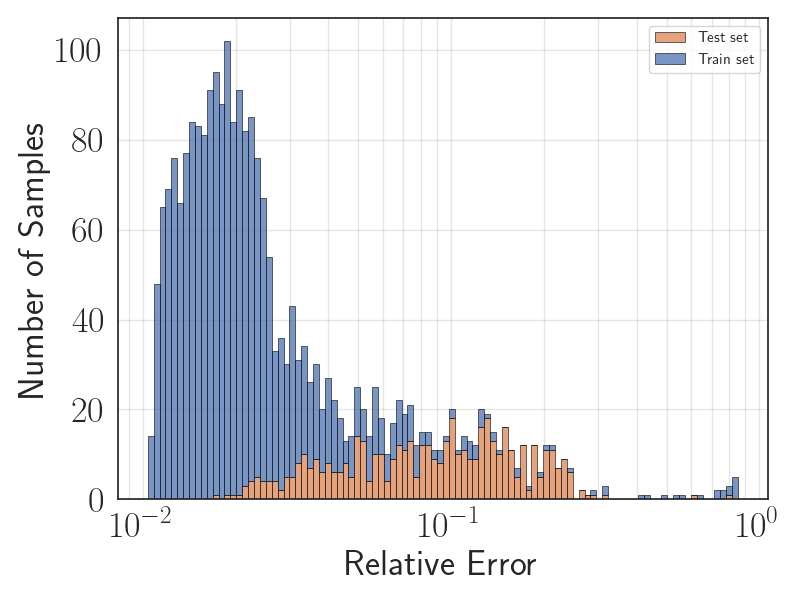}
        \caption{Bar plot for the CNO.}
        \label{fig:barplot_CNO}
    \end{subfigure}%
    \caption{CNO performance metrics: (a) Evolution of the relative $L^2$ training loss (blue), relative $L^1$ test loss (green), and relative $L^2$ test loss (red) across epochs. (b) Distribution of the relative $L^2$ error for the training and test sets.}
    \label{fig:loss_bar_CNO}
\end{figure}

The hyperparameters that are employed for the CNO \cite{CNO23raonic}, as well as the range considered for the automatic hyperparameter tuning, are listed in the Appendix \ref{app_subsec:CNO} in particular, in Table \ref{table:CNO hyperparams}. Figure \ref{fig:loss_bar_CNO} shows the evolution of relative $L^2$ training loss and relative $L^1$ and $L^2$ test losses, averaged over five random seeds (Fig. \ref{fig:loss_CNO}), as well as the bar plots illustrating the distribution of errors for the training and test sets (Fig. \ref{fig:barplot_CNO}).

\begin{figure}[!ht]
    \centering
    \includegraphics[width=\textwidth]{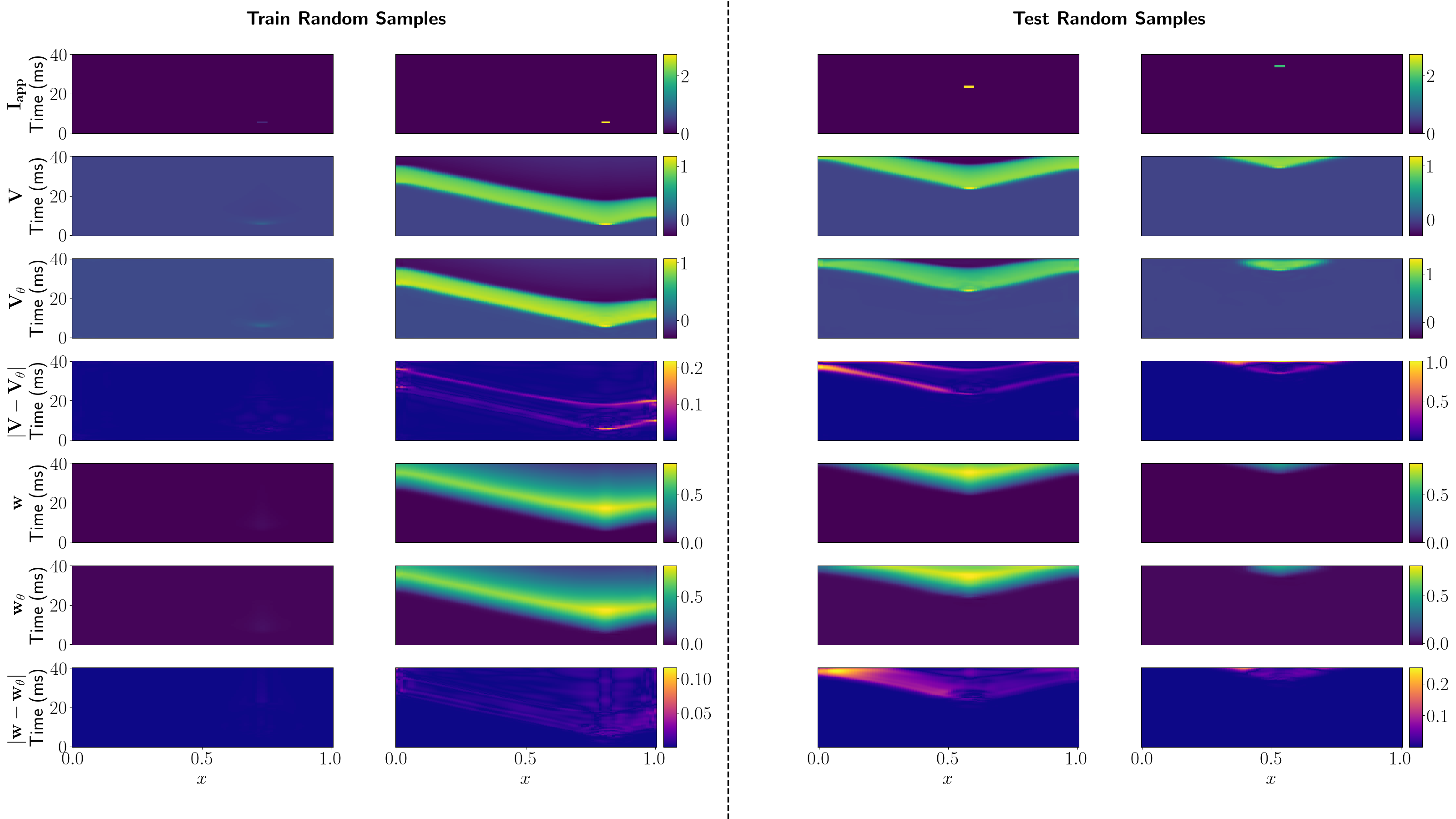}
    \caption{CNO visualization: Four randomly selected examples from the training and test sets are shown. Within each column, the rows illustrate: the applied current $I_{\text{app}}$, high-fidelity reference for potential $V$, CNO prediction $V_{\theta}$ for $V$, pointwise error for $V$, high-fidelity reference for recovery variable $w$, CNO prediction $w_{\theta}$ for $w$, pointwise error $w$.}
    \label{fig:visualization_CNO}
\end{figure}

From the error distribution in Figure \ref{fig:barplot_CNO}, we can observe that CNO performs well in the training with a relative $L^2$ error around $10^{-2}$. However, there are some outliers. Concerning the test set, we observe that we have higher errors. Even considering this, we see that CNOs can capture the translation invariance of the FHN model with slightly less accuracy. Figure \ref{fig:visualization_CNO} shows four randomly selected examples from the training and test sets to visualize the results.

\subsection{Deep Operator Networks results}\label{subsec:DON_results}
\begin{figure}[!ht]
    \begin{subfigure}[t]{0.49\textwidth}
        \centering
        \includegraphics[width=0.9\textwidth]{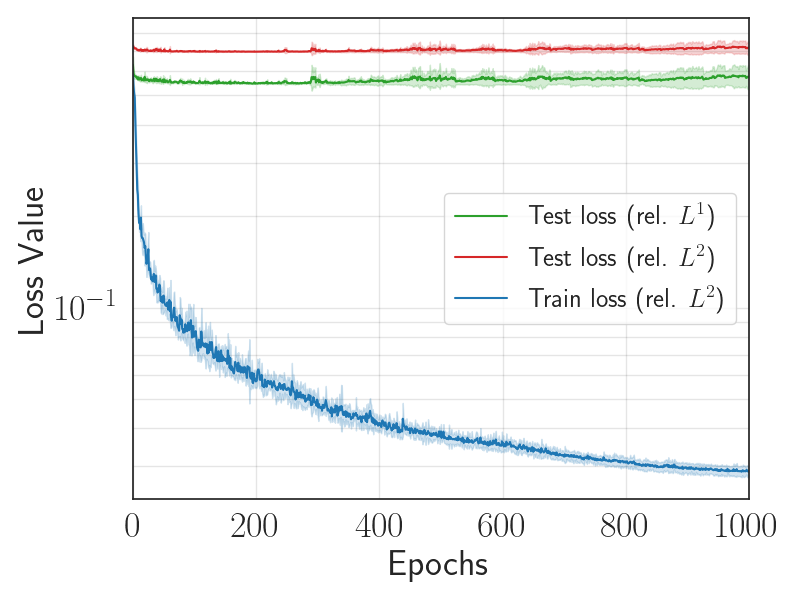}
        \caption{Loss values for the DONs.}
        \label{fig:loss_DON}
    \end{subfigure}%
    \begin{subfigure}[t]{0.49\textwidth}
        \centering
        \includegraphics[width=0.9\textwidth]{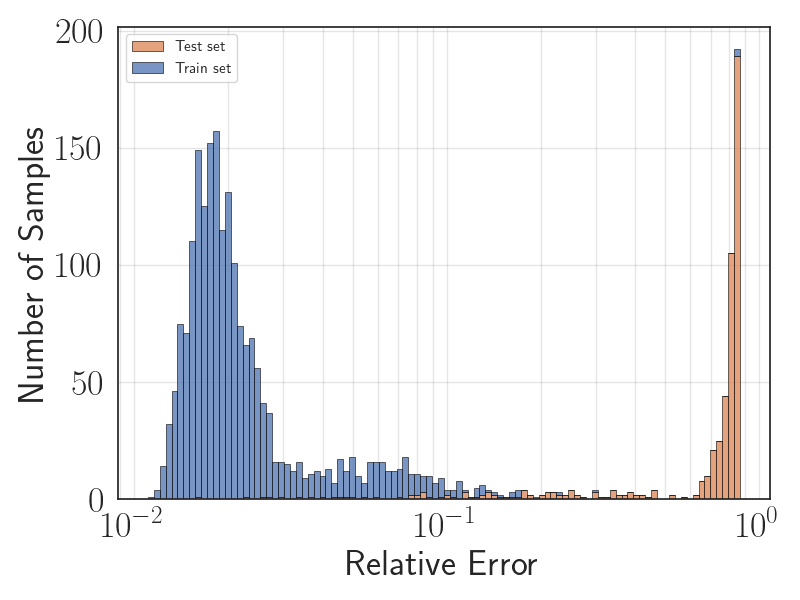}
        \caption{Bar plot for the DONs.}
        \label{fig:barplot_DON}
    \end{subfigure}%
    \caption{DON performance metrics: (a) Evolution of the relative $L^2$ training loss (blue), relative $L^1$ test loss (green), and relative $L^2$ test loss (red) across epochs. (b) Distribution of the relative $L^2$ error for the training and test sets.}
    \label{fig:loss_bar_DON}
\end{figure}

\begin{figure}[!ht]
    \centering
    \includegraphics[width=\textwidth]{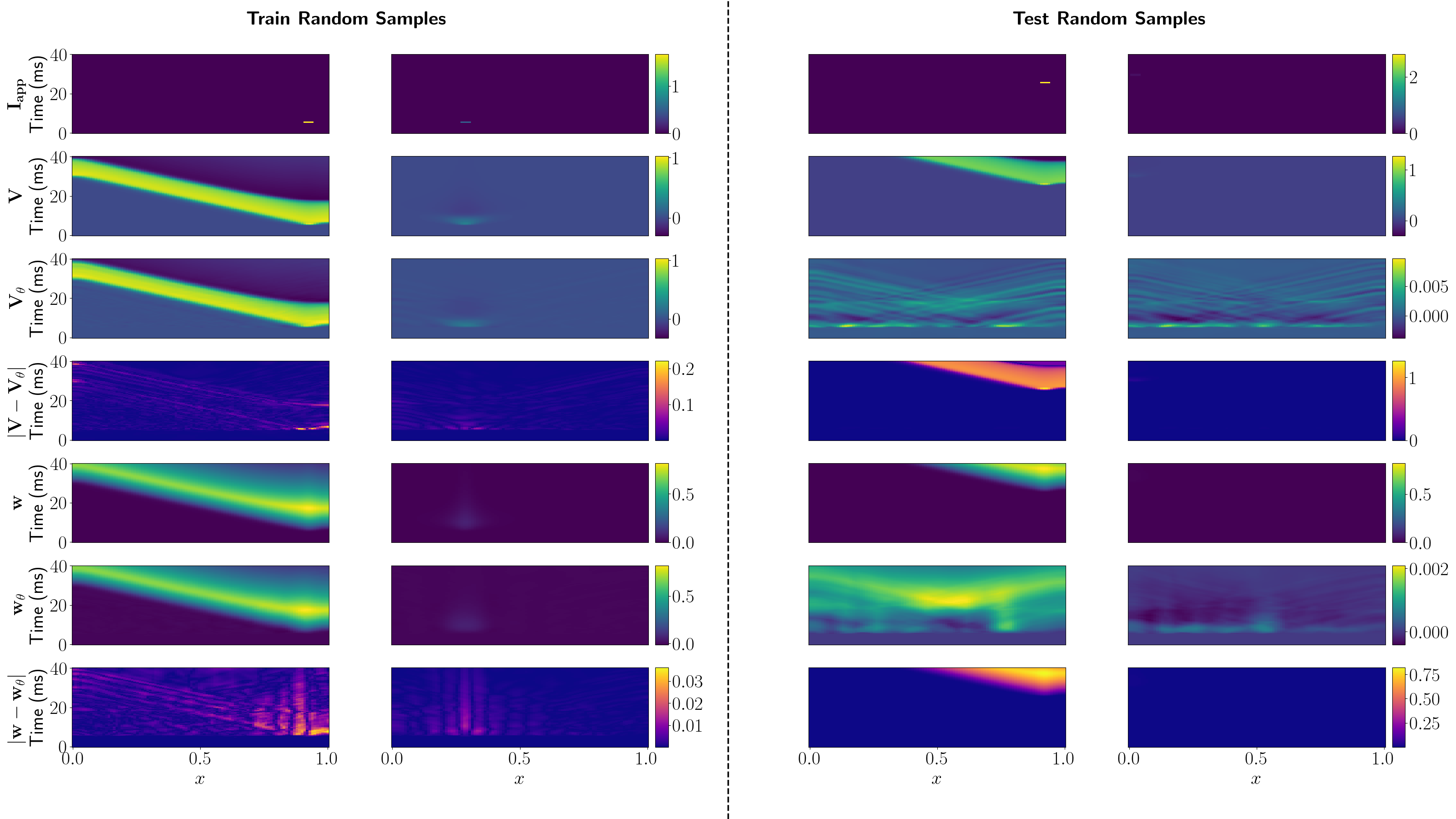}
    \caption{DON visualization: Four randomly selected examples from the training and test sets are shown. Within each column, the rows illustrate: the applied current $I_{\text{app}}$, high-fidelity reference for potential $V$, DON prediction $V_{\theta}$ for $V$, pointwise error for $V$, high-fidelity reference for recovery variable $w$, DON prediction $w_{\theta}$ for $w$, pointwise error $w$.}
    \label{fig:visualization_DON}
\end{figure}

We report the results obtained for the DON \cite{DON21lu} without any modification, i.e., both the branch and the trunk are feed-forward neural networks (FNNs). The hyperparameters that are employed and the range considered for the automatic hyperparameter tuning, are listed in the Appendix \ref{app_subsec:DON} in particular, in Table \ref{table:DON_hyperparams}. Figure \ref{fig:loss_bar_DON} shows the evolution of relative $L^2$ training loss and relative $L^1$ and $L^2$ test losses, averaged over five random seeds (Fig. \ref{fig:loss_DON}), as well as the bar plots illustrating the distribution of errors for the training and test sets (Fig. \ref{fig:barplot_DON}).
From the error distribution in Figure \ref{fig:barplot_DON}, we can observe that DON performs well in the training with a relative $L^2$ error around $10^{-2}$. However, when it comes to the test set, we see that DONs face challenges replicating the FHN translation invariance property. To further highlight this, we visualize the results in Figure \ref{fig:visualization_DON}, which shows four randomly selected examples from the training and test sets.

\subsection{Deep Operator Networks with CNN encoders results}\label{subsec:DON_CNN_results}
\begin{figure}[!ht]
    \begin{subfigure}[t]{0.49\textwidth}
        \centering
        \includegraphics[width=0.9\textwidth]{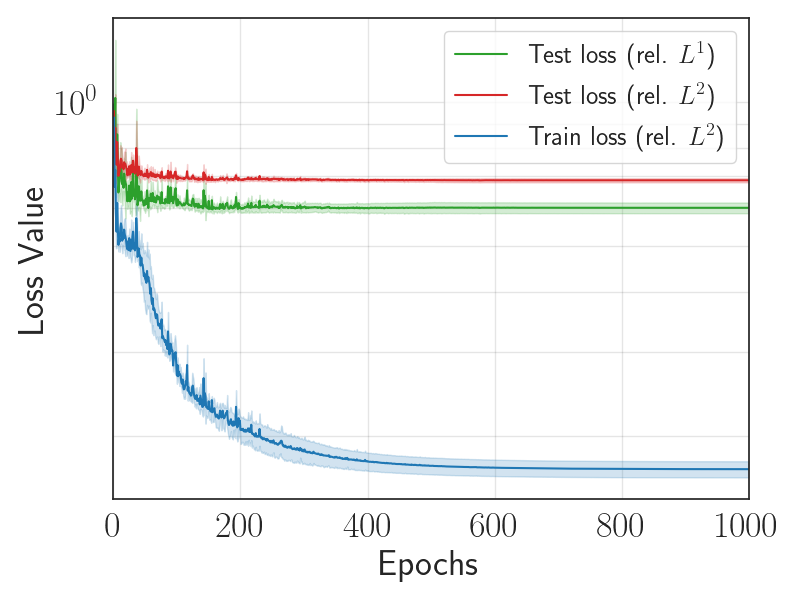}
        \caption{Loss values for the DON-CNN.}
        \label{fig:loss_DON_CNN}
    \end{subfigure}%
    \begin{subfigure}[t]{0.49\textwidth}
        \centering
        \includegraphics[width=0.9\textwidth]{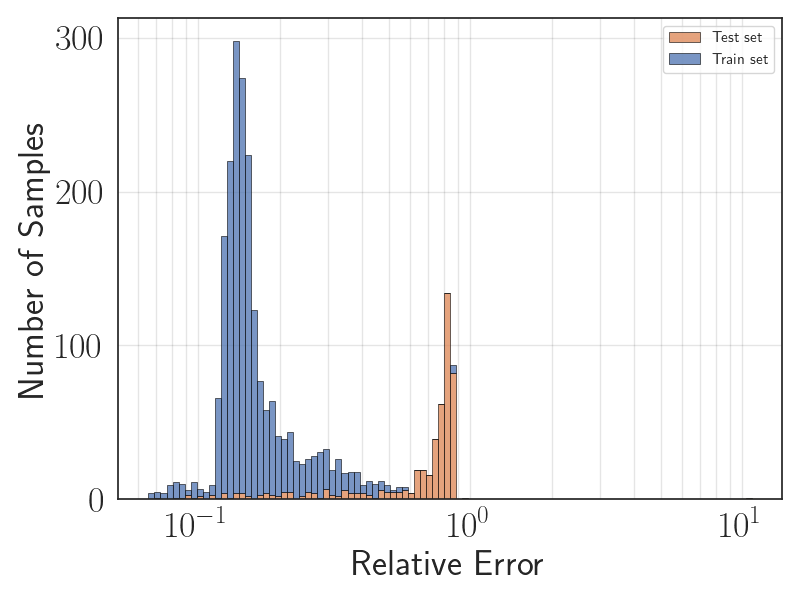}
        \caption{Bar plot for the DON-CNN}
        \label{fig:barplot_DON_CNN}
    \end{subfigure}%
    \caption{DON-CNN performance metrics: (a) Evolution of the relative $L^2$ training loss (blue), relative $L^1$ test loss (green), and relative $L^2$ test loss (red) across epochs. (b) Distribution of the relative $L^2$ error for the training and test sets.}
    \label{fig:loss_bar_DON_CNN}
\end{figure}

We report the results obtained for the DON-CNN i.e., the branch has a CNN encoder before the FNN. The hyperparameters that are employed and the range considered for the automatic hyperparameter tuning, are listed in the Appendix \ref{app_subsec:DON_CNN} in particular, in Table \ref{table:DON_CNN_hyperparams}. Figure \ref{fig:loss_bar_DON_CNN} shows the evolution of relative $L^2$ training loss and relative $L^1$ and $L^2$ test losses, averaged over five random seeds (Fig. \ref{fig:loss_DON_CNN}), as well as the bar plots illustrating the distribution of errors for the training and test sets (Fig. \ref{fig:barplot_DON_CNN}).

\begin{figure}[!ht]
    \centering
    \includegraphics[width=\textwidth]{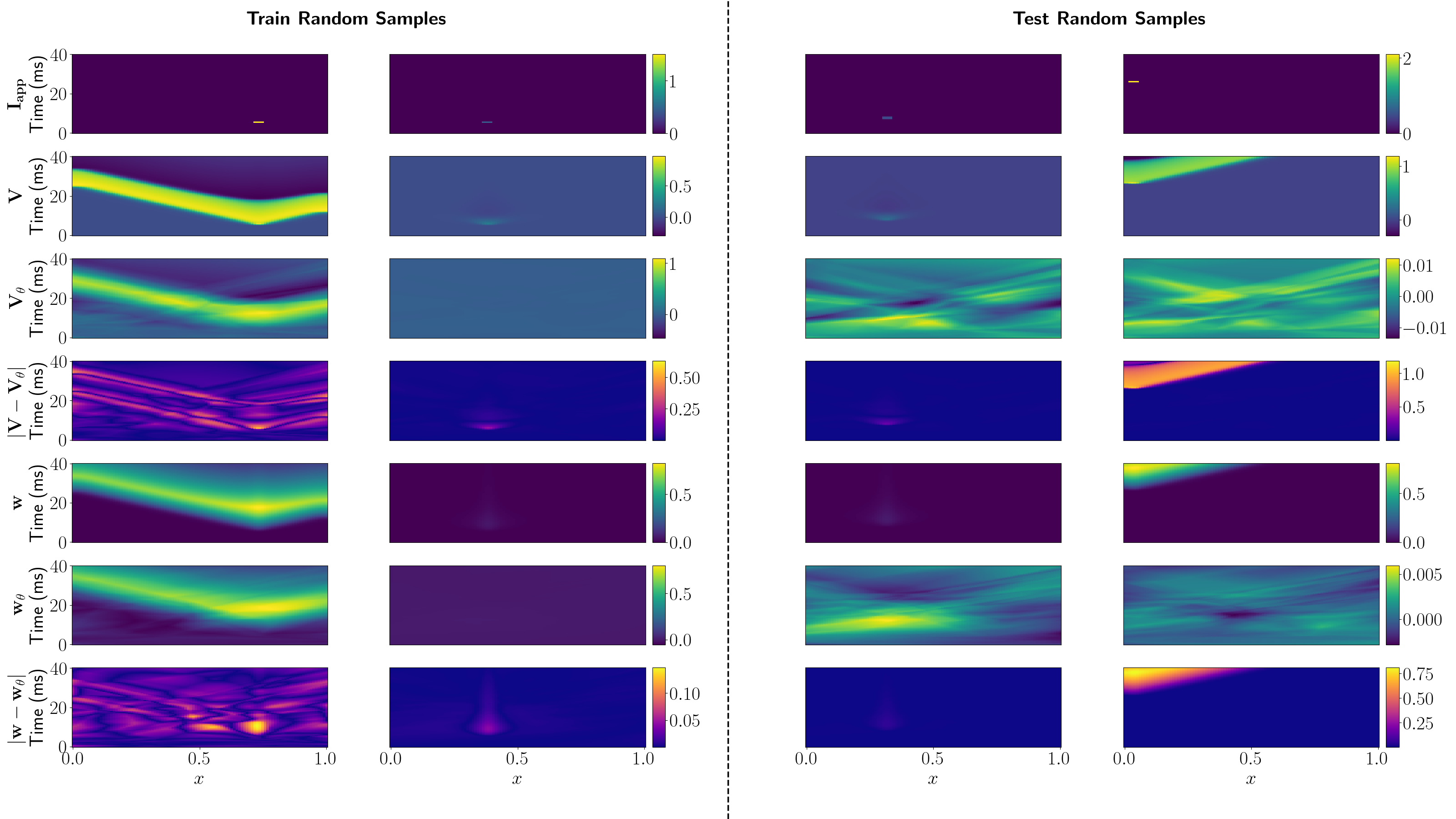}
    \caption{DON-CNN visualization: Four randomly selected examples from the training and test sets are shown. Within each column, the rows illustrate: the applied current $I_{\text{app}}$, high-fidelity reference for potential $V$, DON-CNN prediction $V_{\theta}$ for $V$, pointwise error for $V$, high-fidelity reference for recovery variable $w$, DON-CNN prediction $w_{\theta}$ for $w$, pointwise error $w$.}
    \label{fig:visualization_DON_CNN}
\end{figure}

As shown in Figure \ref{fig:barplot_DON_CNN}, we can observe that DON-CNN can capture the dynamics in the training set with an accuracy of around 10$^{-1}$. However, when it comes to the test set, we see that DONs-CNN face challenges replicating the FHN translation invariance property. Figure \ref{fig:visualization_DON_CNN} shows four randomly selected examples from the training and test sets to visualize the results.

\subsection{Proper Orthogonal Decomposition DONs results}\label{subsec:POD_DON_results}
\begin{figure}[!ht]
    \begin{subfigure}[t]{0.49\textwidth}
        \centering
        \includegraphics[width=0.9\textwidth]{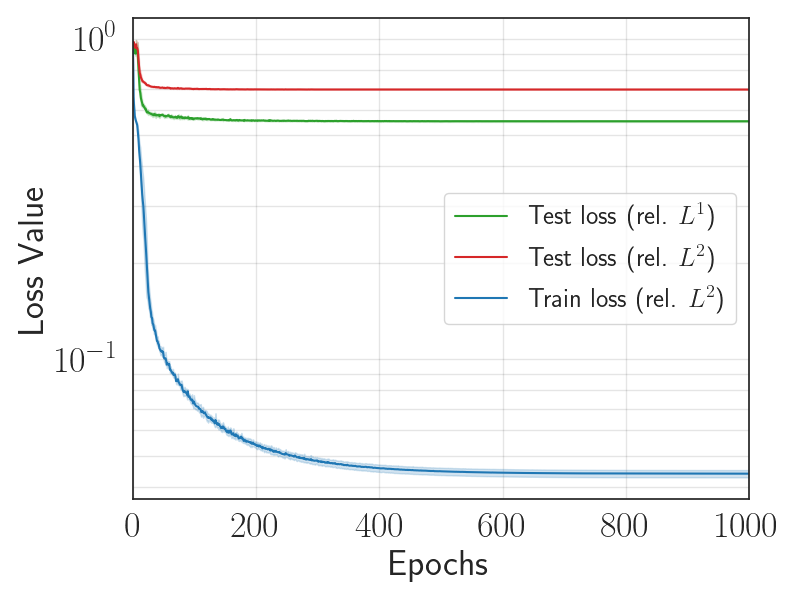}
        \caption{Loss values for the POD-DON.}
        \label{fig:loss_POD_DON}
    \end{subfigure}%
    \begin{subfigure}[t]{0.49\textwidth}
        \centering
        \includegraphics[width=0.9\textwidth]{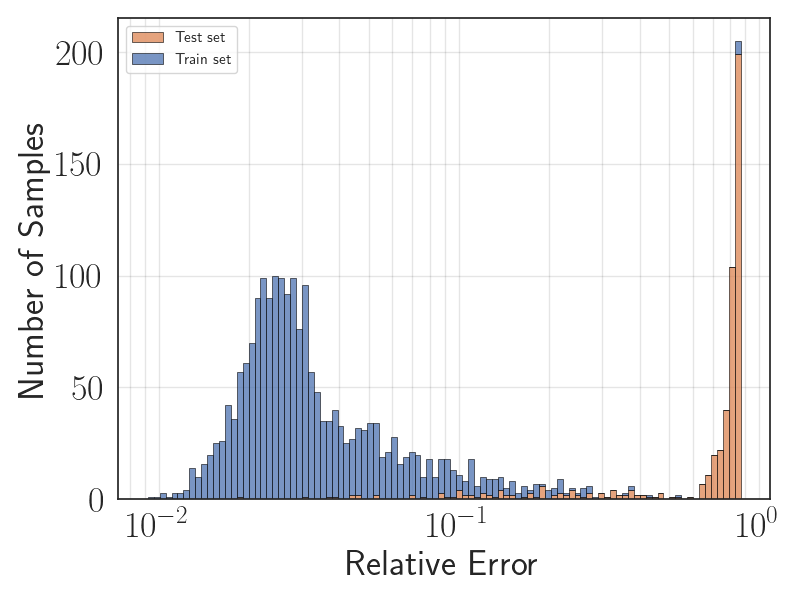}
        \caption{Bar plot for the POD-DON.}
        \label{fig:barplot_POD_DON}
    \end{subfigure}%
    \caption{POD-DON performance metrics: (a) Evolution of the relative $L^2$ training loss (blue), relative $L^1$ test loss (green), and relative $L^2$ test loss (red) across epochs. (b) Distribution of the relative $L^2$ error for the training and test sets.}
    \label{fig:loss_bar_POD_DON}
\end{figure}

We report the results obtained for the POD-DON \cite{lu2022comprehensive} i.e., the trunk is a linear combination of the POD modes. The hyperparameters that are employed and the range considered for the automatic hyperparameter tuning, are listed in the Appendix \ref{app_subsec:POD_DON} in particular, in Table \ref{table:POD_DON_hyperparams}. Figure \ref{fig:loss_bar_POD_DON} shows the evolution of relative $L^2$ training loss and relative $L^1$ and $L^2$ test losses, averaged over five random seeds (Fig. \ref{fig:loss_POD_DON}), as well as the bar plots illustrating the distribution of errors for the training and test sets (Fig. \ref{fig:barplot_POD_DON}).

\begin{figure}[!ht]
    \centering
    \includegraphics[width=\textwidth]{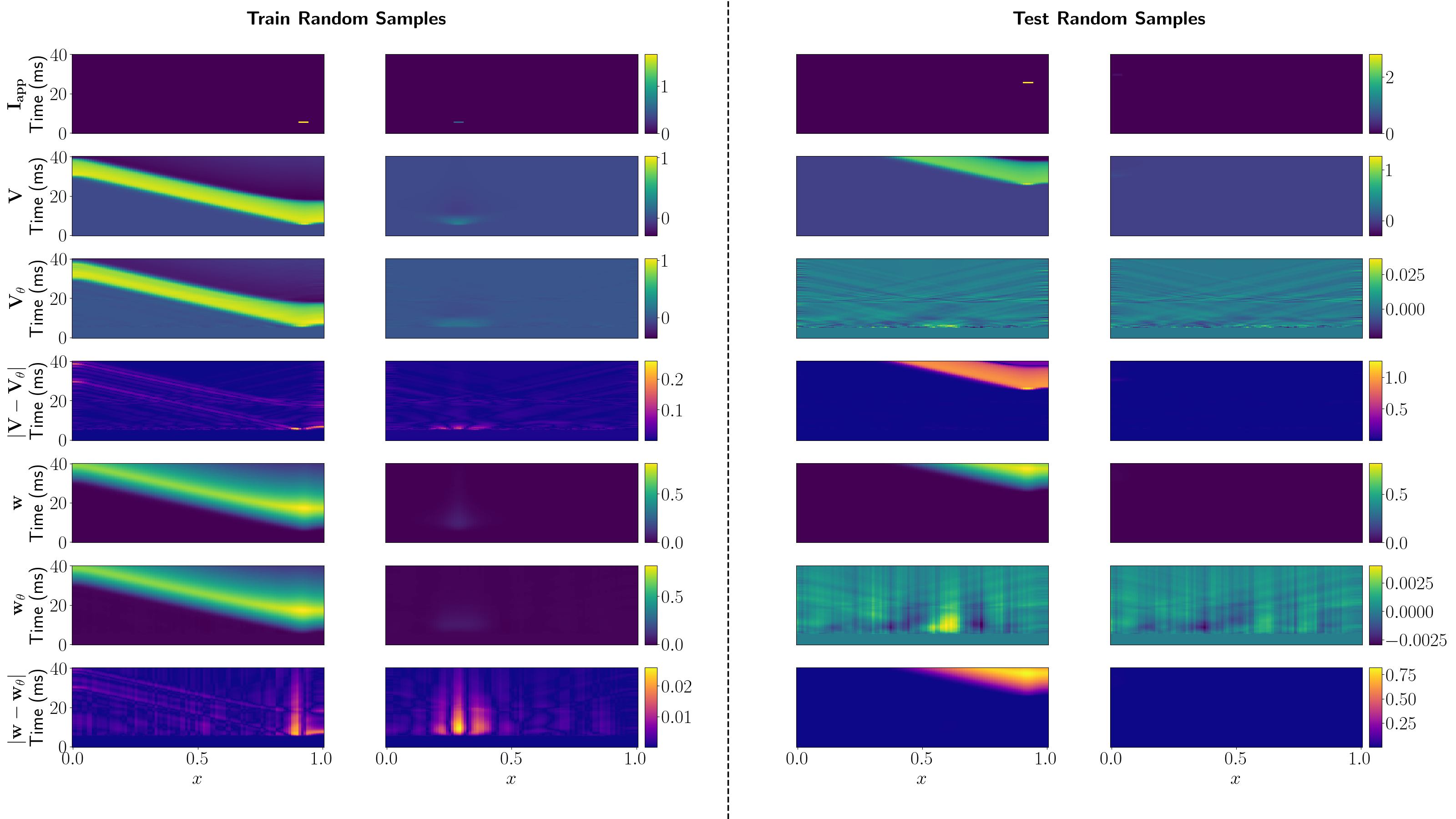}
    \caption{POD-DON visualization: Four randomly selected examples from the training and test sets are shown. Within each column, the rows illustrate: the applied current $I_{\text{app}}$, high-fidelity reference for potential $V$, POD-DON prediction $V_{\theta}$ for $V$, pointwise error for $V$, high-fidelity reference for recovery variable $w$, POD-DON prediction $w_{\theta}$ for $w$, pointwise error $w$.}
    \label{fig:visualization_POD_DON}
\end{figure}

From the error distribution in Figure \ref{fig:barplot_POD_DON}, we can observe that POD-DON achieves a relative $L^2$ error around $0.05$ on the training set. However, when it comes to the test set, we see that POD-DONs face challenges replicating the FHN translation invariance property. Figure \ref{fig:visualization_POD_DON} shows four randomly selected examples from the training and test sets to visualize the results.

\clearpage

\subsection{Fourier Neural Operators}\label{subsec:FNO_results}
\begin{figure}[!ht]
    \begin{subfigure}[t]{0.49\textwidth}
        \centering
        \includegraphics[width=0.9\textwidth]{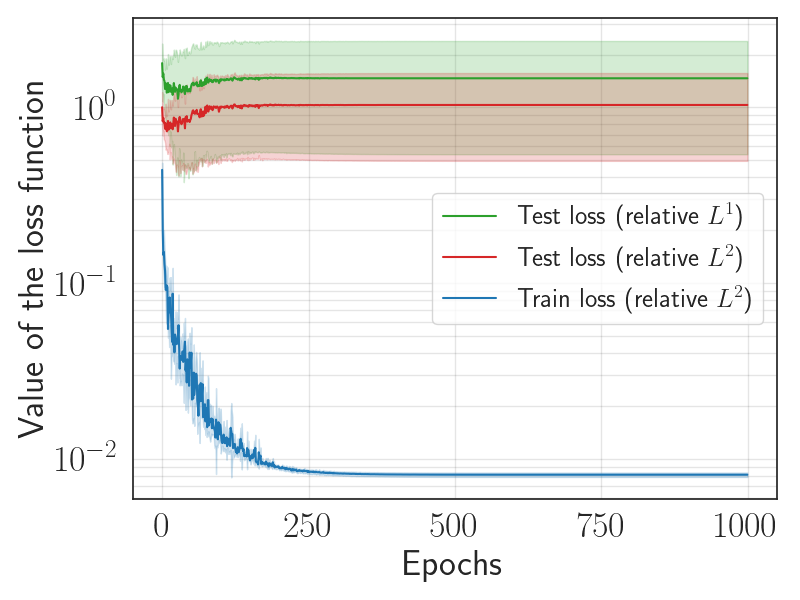}
        \caption{Loss values for the FNO.}
        \label{fig:loss_FNO}
    \end{subfigure}%
    \begin{subfigure}[t]{0.49\textwidth}
        \centering
        \includegraphics[width=0.9\textwidth]{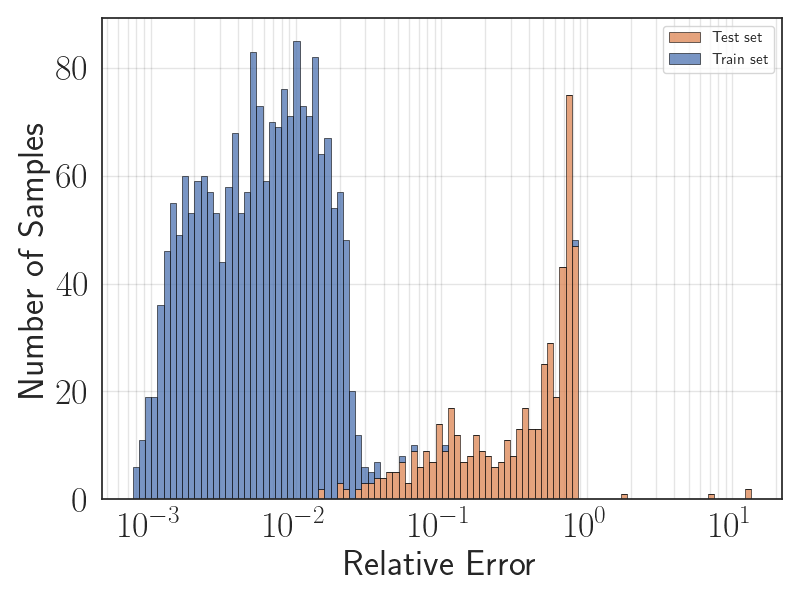}
        \caption{Bar plot for the FNO.}
        \label{fig:barplot_FNO}
    \end{subfigure}%
    \caption{FNO performance metrics: (a) Evolution of the relative $L^2$ training loss (blue), relative $L^1$ test loss (green), and relative $L^2$ test loss (red) across epochs. (b) Distribution of the relative $L^2$ error for the training and test sets.}
    \label{fig:loss_bar_FNO}
\end{figure}

We report the results obtained for the FNO \cite{FNO20li}. The hyperparameters that are employed and the range considered for the automatic hyperparameter tuning, are listed in the Appendix \ref{app_subsec:FNO} in particular, in Table \ref{table:FNO_hyperparams}.  Figure \ref{fig:loss_bar_FNO} shows the evolution of relative $L^2$ training loss and relative $L^1$ and $L^2$ test losses, averaged over five random seeds (Fig. \ref{fig:loss_FNO}), as well as the bar plots illustrating the distribution of errors for the training and test sets (Fig. \ref{fig:barplot_FNO}).

\begin{figure}[!ht]
    \centering
    \includegraphics[width=\textwidth]{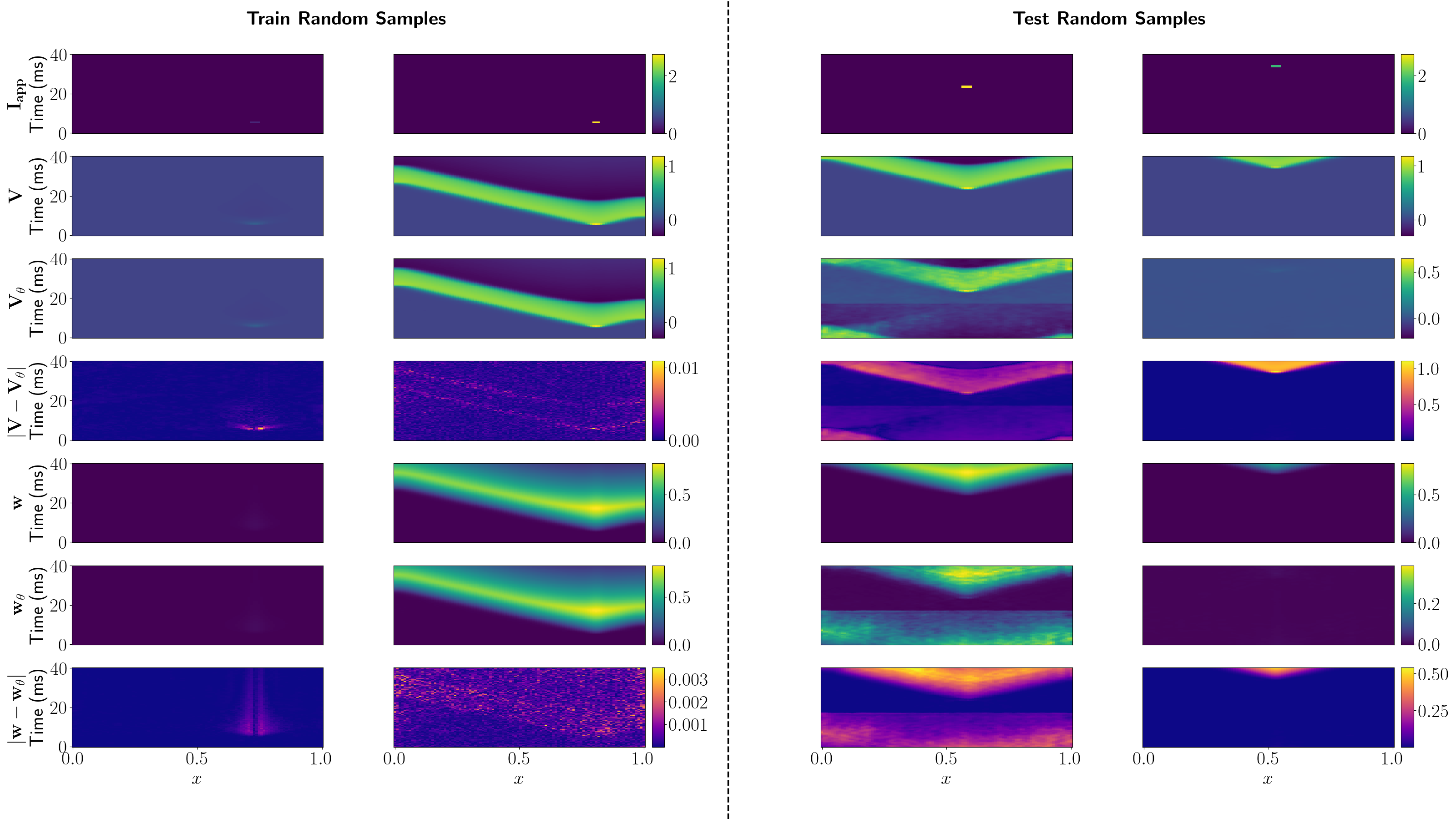}
    \caption{FNO visualization: Five test examples are shown, three randomly selected, and the two with the highest relative $L^2$ error. Within each column, the rows illustrate: the applied current $I_{\text{app}}$, high-fidelity reference for potential $V$, FNO prediction $V_{\theta}$ for $V$, pointwise error for $V$, high-fidelity reference for recovery variable $w$, FNO prediction $w_{\theta}$ for $w$, pointwise error $w$.}
    \label{fig:visualization_FNO}
\end{figure}

The error distribution in Figure \ref{fig:barplot_FNO} indicates that the FNO performs well on the training set, achieving errors as low as $10^{-3}$. However, the FNOs successfully reproduces some cases on the test set. Still, it faces challenges in reliably capturing the model's translation invariance property. In fact, many examples have errors around $10^0$. Furthermore, the loss evolution in Figure \ref{fig:loss_FNO} shows a non-negligible standard deviation for the test loss, indicating sensitivity to initialization. Figure \ref{fig:visualization_FNO} shows four randomly selected examples from the training and test sets to visualize the results.

\subsection{Tucker Tensorized FNOs}\label{subsec:TFNO}

\begin{figure}[!ht]
    \begin{subfigure}[t]{0.49\textwidth}
        \centering
        \includegraphics[width=0.9\textwidth]{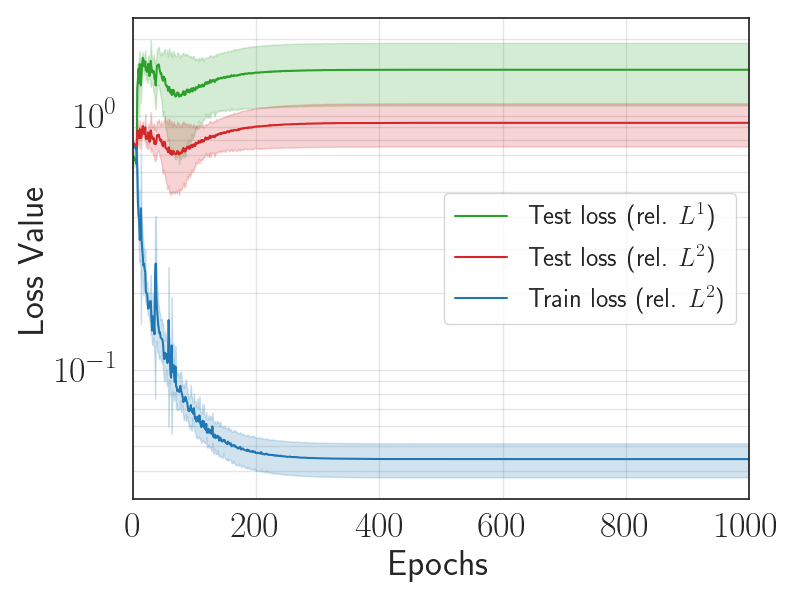}
        \caption{Loss values for the TFNO.}
        \label{fig:loss_TFNO}
    \end{subfigure}%
    \begin{subfigure}[t]{0.49\textwidth}
        \centering
        \includegraphics[width=0.9\textwidth]{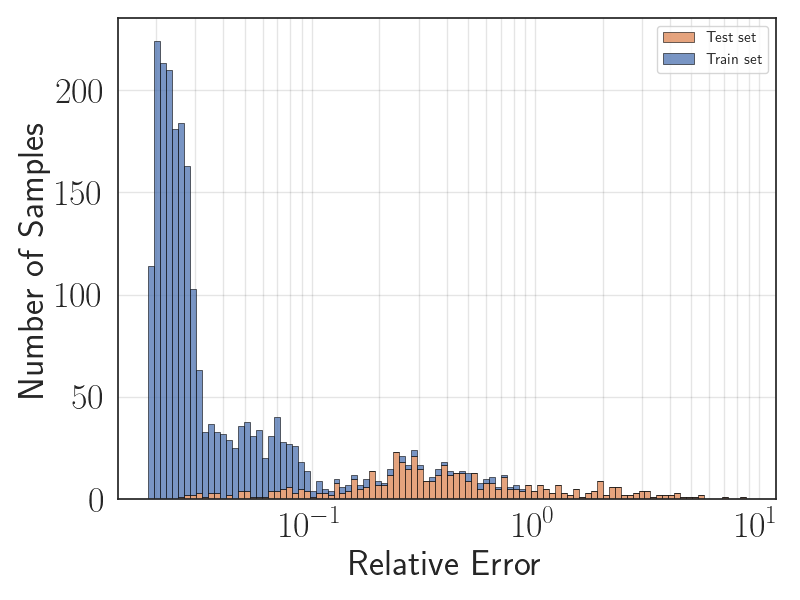}
        \caption{Bar plot for the TFNO.}
        \label{fig:barplot_TFNO}
    \end{subfigure}%
    \caption{TFNO performance metrics: (a) Evolution of the relative $L^2$ training loss (blue), relative $L^1$ test loss (green), and relative $L^2$ test loss (red) across epochs. (b) Distribution of the relative $L^2$ error for the training and test sets.}
    \label{fig:loss_bar_TFNO}
\end{figure}

We report the results obtained for the TFNO \cite{zhou2026tuckerfno}. The hyperparameters that are employed and the range considered for the automatic hyperparameter tuning, are listed in the Appendix \ref{app_subsec:TFNO} in particular, in Table \ref{table:TFNO_hyperparams}.  Figure \ref{fig:loss_bar_TFNO} shows the evolution of relative $L^2$ training loss and relative $L^1$ and $L^2$ test losses, averaged over five random seeds (Fig. \ref{fig:loss_TFNO}), as well as the bar plots illustrating the distribution of errors for the training and test sets (Fig. \ref{fig:barplot_TFNO}).

\begin{figure}[!ht]
    \centering
    \includegraphics[width=\textwidth]{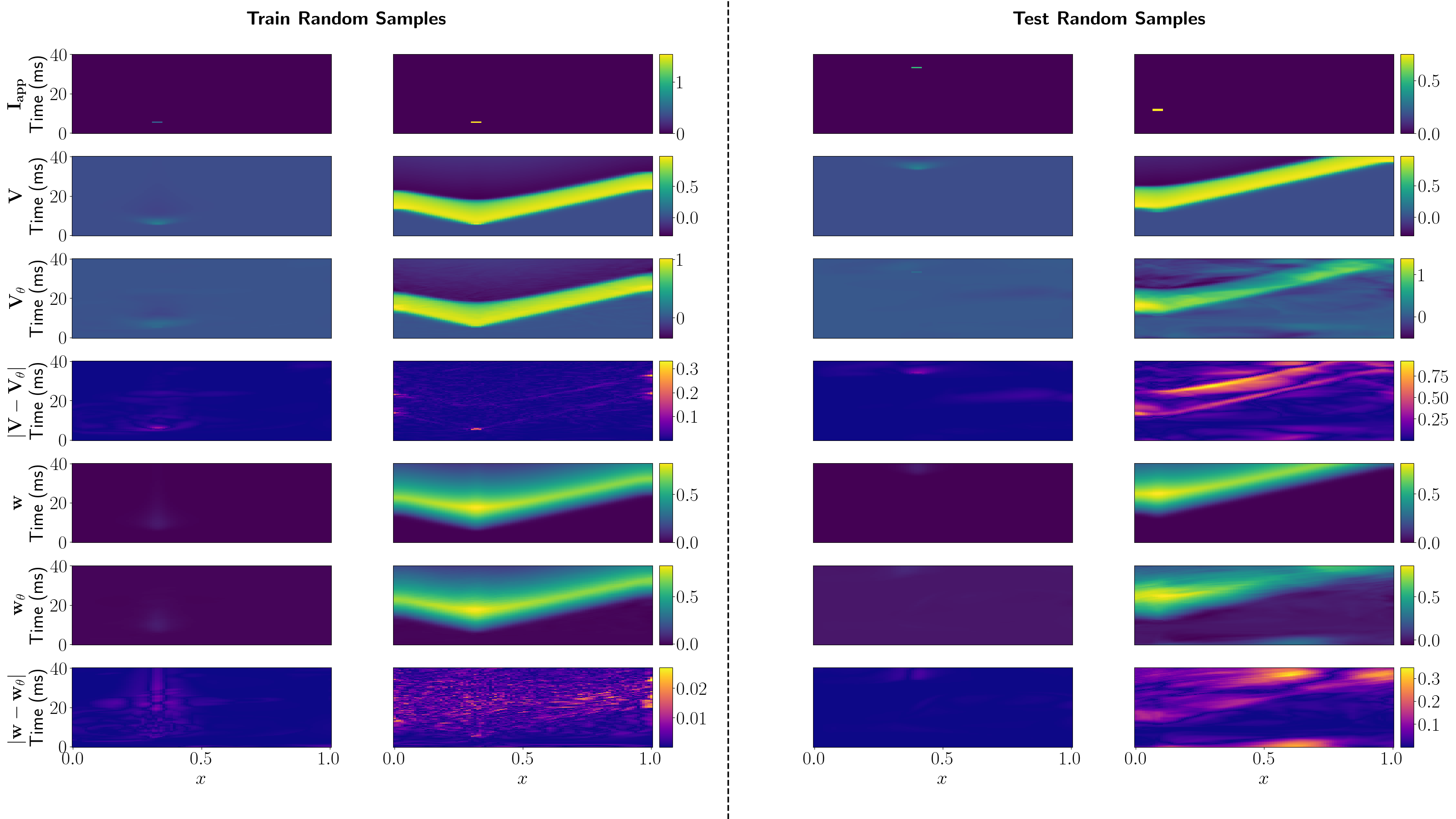}
    \caption{TFNO visualization: Four randomly selected examples from the training and test sets are shown. Within each column, the rows illustrate: the applied current $I_{\text{app}}$, high-fidelity reference for potential $V$, TFNO prediction $V_{\theta}$ for $V$, pointwise error for $V$, high-fidelity reference for recovery variable $w$, TFNO prediction $w_{\theta}$ for $w$, pointwise error $w$.}
    \label{fig:visualization_TFNO}
\end{figure}

From the error distribution in Figure \ref{fig:barplot_TFNO}, we can observe that TFNO achieves a relative $L^2$ error around $0.05$ on the training set. However, on the test set, the TFNOs successfully reproduces some cases, yet it still faces challenges in reliably capturing the model's translation invariance property. In fact, there is the presence of examples with errors bigger than $10^0$. Furthermore, the loss evolution in Figure \ref{fig:loss_TFNO} shows a non-negligible standard deviation for the test loss, indicating sensitivity to initialization. Figure \ref{fig:visualization_TFNO} shows four randomly selected examples from the training and test sets to visualize the results.

\subsection{Local Neural Operators}\label{subsec:LocalNO_resutls}
\begin{figure}[!ht]
    \begin{subfigure}[t]{0.49\textwidth}
        \centering
        \includegraphics[width=0.9\textwidth]{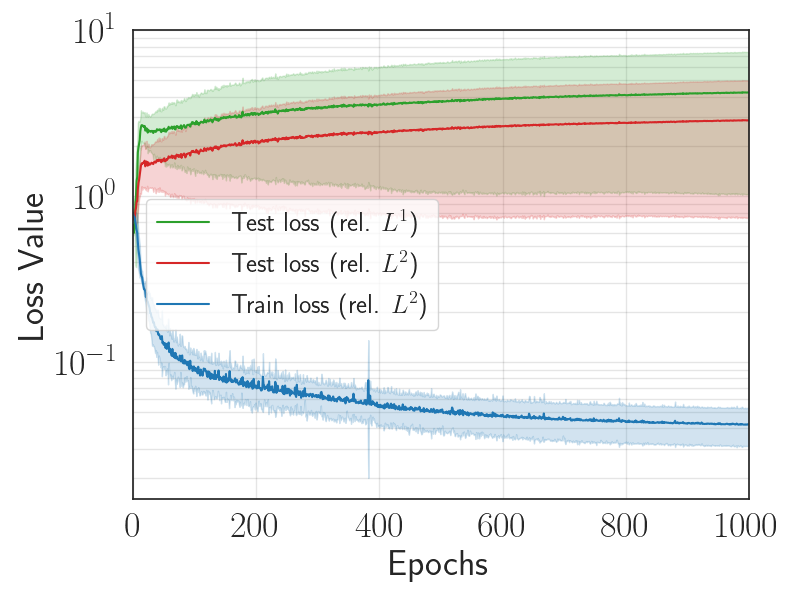}
        \caption{Loss values for the LocalNO.}
        \label{fig:loss_LocalNO}
    \end{subfigure}%
    \begin{subfigure}[t]{0.49\textwidth}
        \centering
        \includegraphics[width=0.9\textwidth]{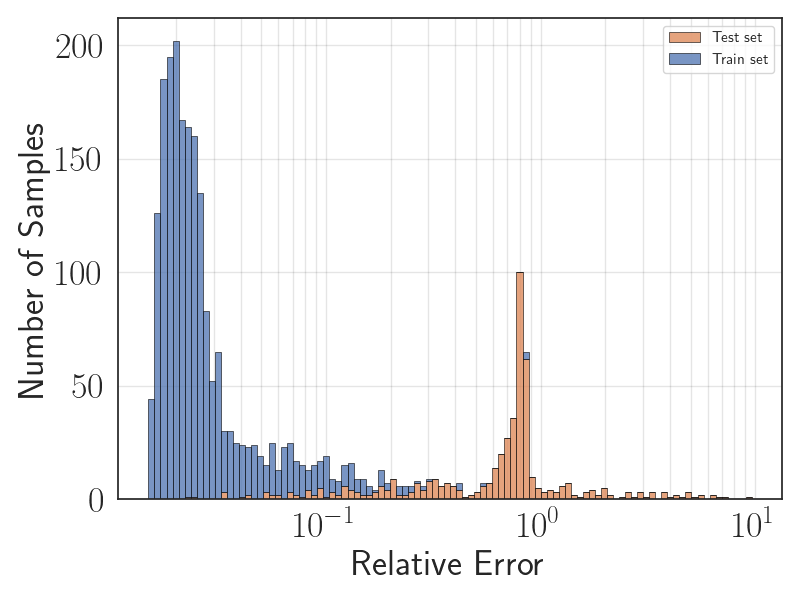}
        \caption{Bar plot for the LocalNO.}
        \label{fig:barplot_LocalNO}
    \end{subfigure}%
    \caption{LocalNO performance metrics: (a) Evolution of the relative $L^2$ training loss (blue), relative $L^1$ test loss (green), and relative $L^2$ test loss (red) across epochs. (b) Distribution of the relative $L^2$ error for the training and test sets.}
    \label{fig:loss_bar_LocalNO}
\end{figure}

We report the results obtained for the LocalNO \cite{liu2024neural}. The hyperparameters that are employed and the range considered for the automatic hyperparameter tuning, are listed in the Appendix \ref{app_subsec:LocalNO} in particular, in Table \ref{table:LocalNO_hyperparams}.  Figure \ref{fig:loss_bar_LocalNO} shows the evolution of relative $L^2$ training loss and relative $L^1$ and $L^2$ test losses, averaged over five random seeds (Fig. \ref{fig:loss_LocalNO}), as well as the bar plots illustrating the distribution of errors for the training and test sets (Fig. \ref{fig:barplot_LocalNO}).

\begin{figure}[!ht]
    \centering
    \includegraphics[width=\textwidth]{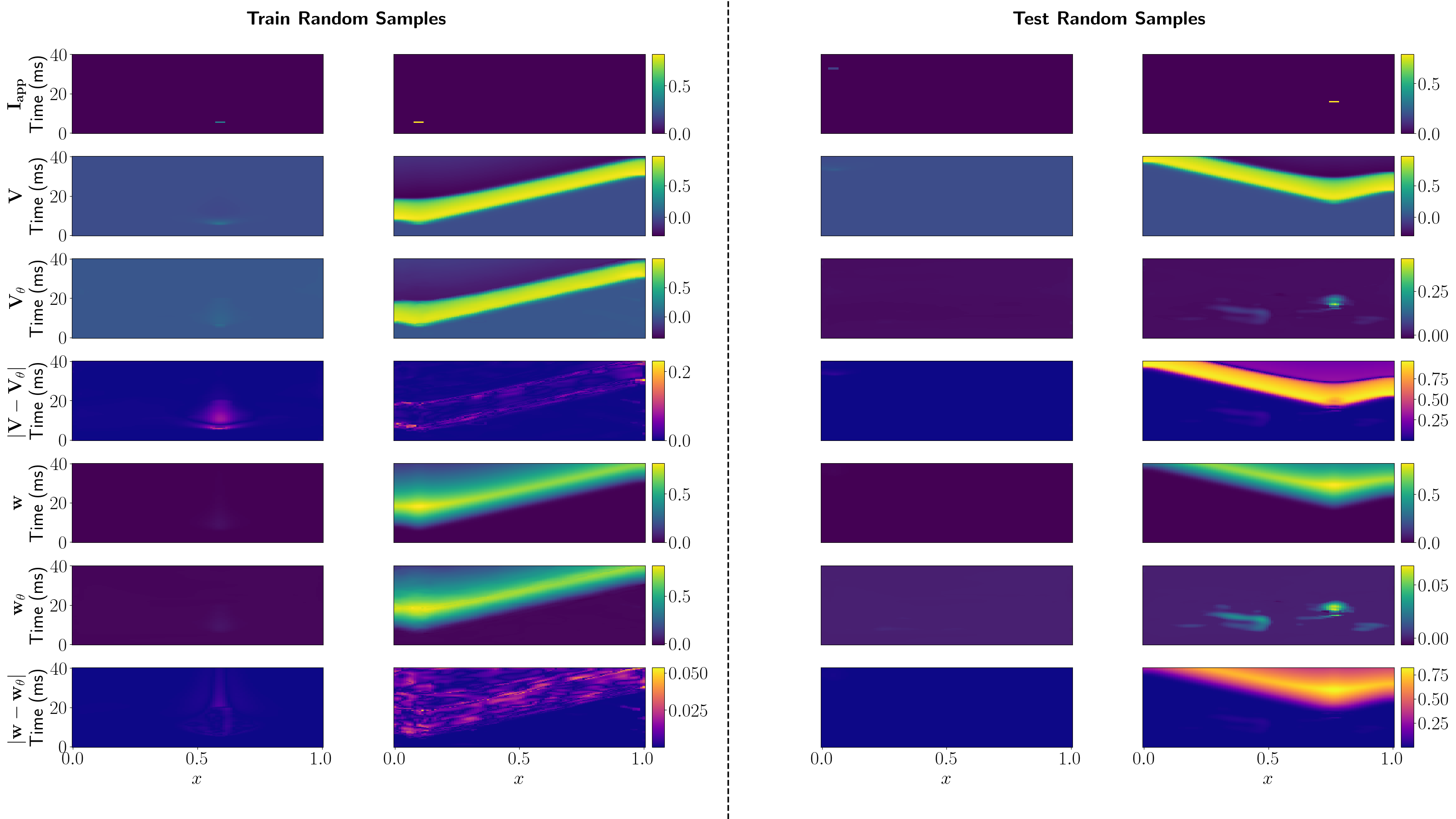}
    \caption{LocalNO visualization: Four randomly selected examples from the training and test sets are shown. Within each column, the rows illustrate: the applied current $I_{\text{app}}$, high-fidelity reference for potential $V$, LocalNO prediction $V_{\theta}$ for $V$, pointwise error for $V$, high-fidelity reference for recovery variable $w$, LocalNO prediction $w_{\theta}$ for $w$, pointwise error $w$.}
    \label{fig:visualization_LocalNO}
\end{figure}

From the error distribution in Figure \ref{fig:barplot_LocalNO}, we can observe that LocalNO achieves a relative $L^2$ error around $0.05$ on the training set. However, on the test set, the LocalNOs successfully reproduces some cases, yet it still faces challenges in reliably capturing the model's translation invariance property. In fact, there is the presence of examples with errors bigger than $10^0$. Furthermore, the loss evolution in Figure \ref{fig:loss_LocalNO} shows a non-negligible standard deviation for the test loss, indicating sensitivity to initialization. Figure \ref{fig:visualization_LocalNO} shows four randomly selected examples from the training and test sets to visualize the results.

\clearpage
\section{Comparison}\label{sec:Comparision}
The following section provides a detailed comparison of the results obtained by the different architectures.  We will compare them in terms of training and test errors, training time, and the number of trainable parameters. Additionally, we will compare the error and inference time for a single example from the test set. In the following Table \ref{table:results} we report the results obtained by the different architectures in terms of the number of trainable parameters, training time, training error and test error. The training time is reported as both the total time required for training and the average time per epoch. Training and test errors are reported in terms of $L^2$ and $L^1$ relative norms, with respective means and standard deviations across five runs with different random seeds.

\begin{table}[ht!]
    \centering
    \resizebox{\textwidth}{!}{
        \begin{tabular}{ccc|cc|cc|cccc}
            \hline
            Architecture & Number of              & \multicolumn{2}{c}{Training Time} & Training Error      & \multicolumn{4}{c}{Test}                                                                                                     \\
                         & parameters             & Total                             & Epoch               & $L^2$                    & \multicolumn{2}{c}{$L^2$} & \multicolumn{2}{c}{$L^1$}                                             \\
                         &                        &                                   &                     &                          & Mean                      & Median                    & Mean                & Median              \\
            \hline
            CNO          & 8.2 M                  & 10 h                              & 36.4 sec            & $0.0323 \pm 0.0059$      & $0.1501 \pm 0.0443$
                         & $0.0985 \pm 0.0152$    & $0.0941 \pm 0.0332$               & $0.0535 \pm 0.0123$                                                                                                                                \\

            DON          & 1.5M                   & 30m                               & 1.73 sec            & $0.0273 \pm 0.0041$      & $0.7190\pm 0.0380$        & $0.8161 \pm 0.0001$       & $0.5756\pm 0.0536$  & $	0.6532\pm 0.0132$ \\
            DON-CNN      & 0.5M                   & 15 min                            & 0.88 sec            & $0.1701 \pm 0.0066$      & $0.6861 \pm 0.0068$       & $0.7772 \pm 0.0178$       & $0.6008\pm0.0155$   & $	0.6559\pm0.0241$  \\
            POD-DON      & 1M                     & 16m                               & 0.98 sec            & $0.0439 \pm 0.0012$      & $0.6953\pm 0.0007$        & $	0.8146\pm 0.0009$       & $	0.5535\pm 0.0028$ & $0.6497 \pm 0.0010$ \\
            FNO          & 151.1  M               & 4 h                               & 15.8 sec            & $0.0081 \pm 0.0003$
                         & $1.0333    \pm 0.6000$ & $0.5393 \pm0.1218 $               & $1.4655\pm 1.0330$  & $ 0.4894 \pm 0.1041$                                                                                                         \\
            LocalNO      & 0.15 M                 & 1h                                & 4.91 sec            & $0.0423 \pm 0.0121$
                         & $2.8717    \pm 2.3716$ & $1.4850\pm 1.4840 $               & $4.2232 \pm 3.5578$ & $ 1.5667 \pm 1.6579$                                                                                                         \\
            TFNO         & 0.3 M                  & 9h                                & 35.95 sec           & $0.0444\pm0.0067$
                         & $0.9378   \pm 0.1813$  & $0.4378\pm 0.1259 $               & $1.5174\pm 0.4191$  & $0.4070 \pm 0.0905$                                                                                                          \\
            \hline
        \end{tabular}}
    \caption{Summary of the results obtained by the different architecture.}
    \label{table:results}
\end{table}  

\begin{figure}[!ht]
    \centering
    \includegraphics[width=\textwidth]{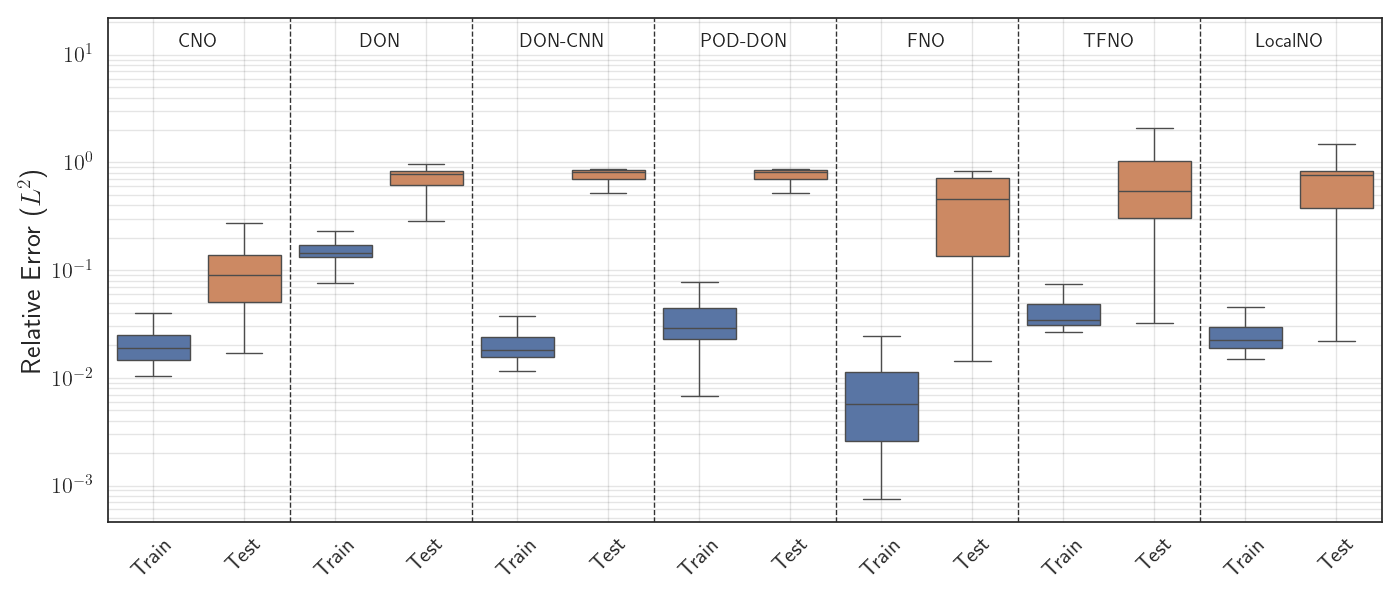}
    \caption{Box plots illustrating the distribution of relative $L^2$ errors for both the training and test sets for all architectures taken into consideration.}
    \label{fig:multiple box plot}
\end{figure}

As illustrated in Table \ref{table:results}, there is a significant discrepancy in the number of trainable parameters across the architectures. This variation was not manually controlled but was determined by the automated hyperparameter tuning process, with more details provided in Appendix \ref{app_sec:ray}. Each architecture was independently optimized over 100 trials on the validation set. We now shift our focus to the performance distribution. Figure \ref{fig:multiple box plot} presents the box plots for training and test errors.

As illustrated in Figure \ref{fig:multiple box plot}, FNO achieves the highest accuracy in the training set. The other architectures exhibit comparable performance, with the exception of DON, which yields a higher error. However, a different trend emerges in the test set, where the models must account for the translated dynamics. Here, only CNO demonstrates the ability to generalize effectively. In contrast, DON and its variants (DON-CNN and POD-DON) face challenges in replicating the model’s translation invariance property (see Figs. \ref{fig:visualization_DON}, \ref{fig:visualization_DON_CNN}, and \ref{fig:visualization_POD_DON}). While FNOs and their variants (TFNOs and LocalNOs) can produce accurate results for certain test examples, there are many examples with high $L^2$ relative errors (see Figs. \ref{fig:barplot_FNO}, \ref{fig:barplot_TFNO}, and \ref{fig:barplot_LocalNO}). This comparison suggests a trade-off: while FNO is the optimal choice for the training set in terms of accuracy, CNO provides a more reliable and robust solution for out-of-distribution tasks involving translations. However, accuracy alone is not a comprehensive evaluation of deep learning methods.  To provide a more comprehensive comparison, we must also account for computational efficiency, specifically training time and the number of trainable parameters. To this end, we evaluate the following two metrics:

\[\]
\[
    \begin{split}
         & \mathcal{C} \coloneqq E \times P \times T                                                                                                         \\
         & \mathcal{C}_{log} \coloneqq E \times \Big[1 + \alpha \log_{10}(\frac{P}{P_{min}})\Big] \times  \Big[1 + \beta \log_{10}(\frac{T}{T_{min}})\Big] \\
 \end{split}
\]

\begin{figure}[!ht]
    \centering
    \includegraphics[width=\textwidth]{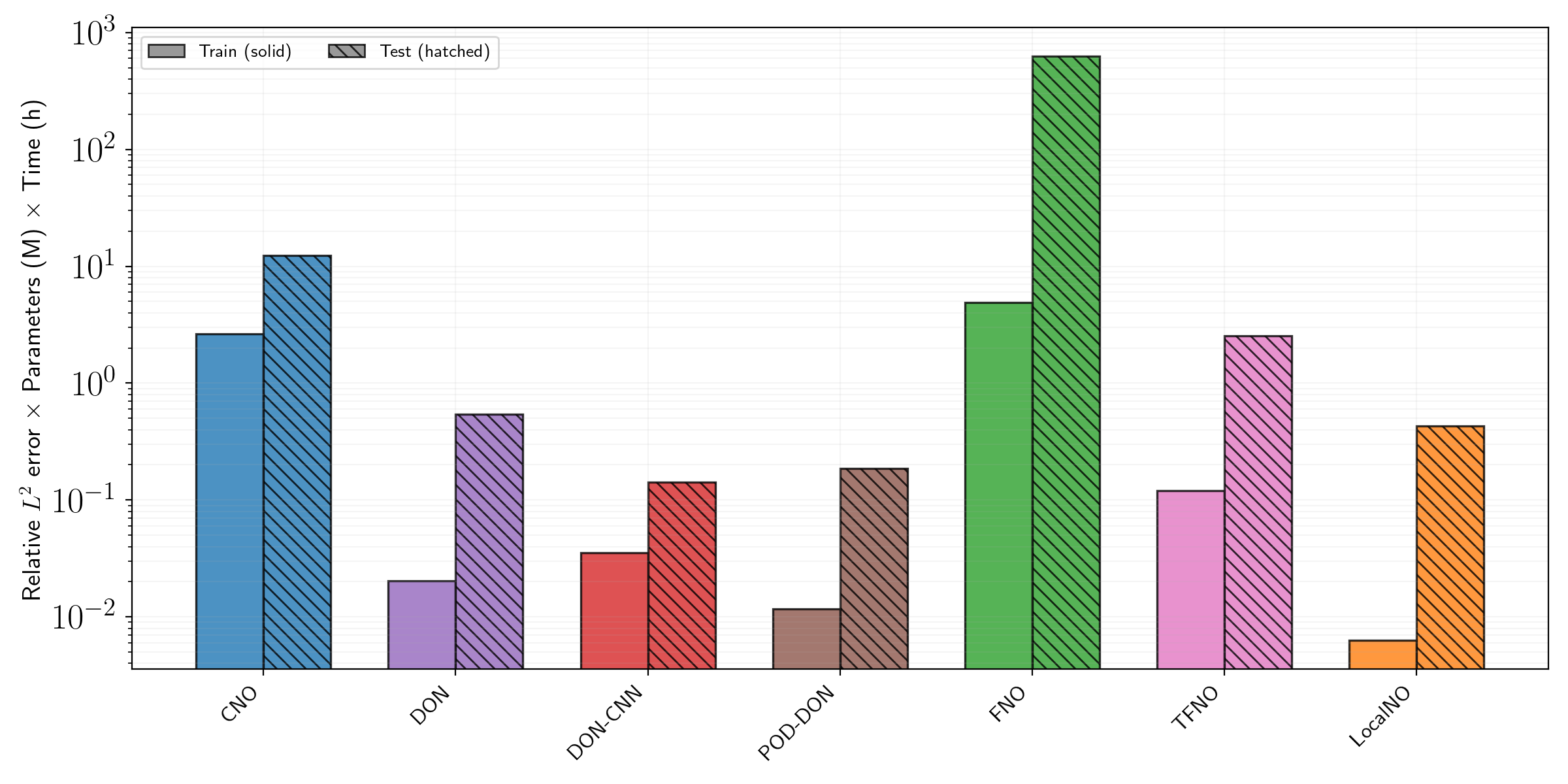}
    \caption{Histogram of the $\mathcal{C}$ cost values for each architecture. The cost related to the training error is shown in solid color, while the test is indicated by the hatched pattern.}
    \label{fig:Cost}
\end{figure}

Where $E$ denotes the relative $L^2$ error, $P$ is the number of trainable parameters (in millions), and $T$ represents the total training time in hours. The terms $P_{\min}$ and $T_{\min}$ correspond to the minimum number of parameters and the shortest training time, respectively, among all architectures reported in Table \ref{table:results}. The weighting coefficients $\alpha$ and $\beta$ are both fixed at $0.5$. Figure \ref{fig:Cost}  shows the cost function $\mathcal{C}$ values for training and test errors of each architecture. 

\begin{figure}[!ht]
    \centering
    \includegraphics[width=\textwidth]{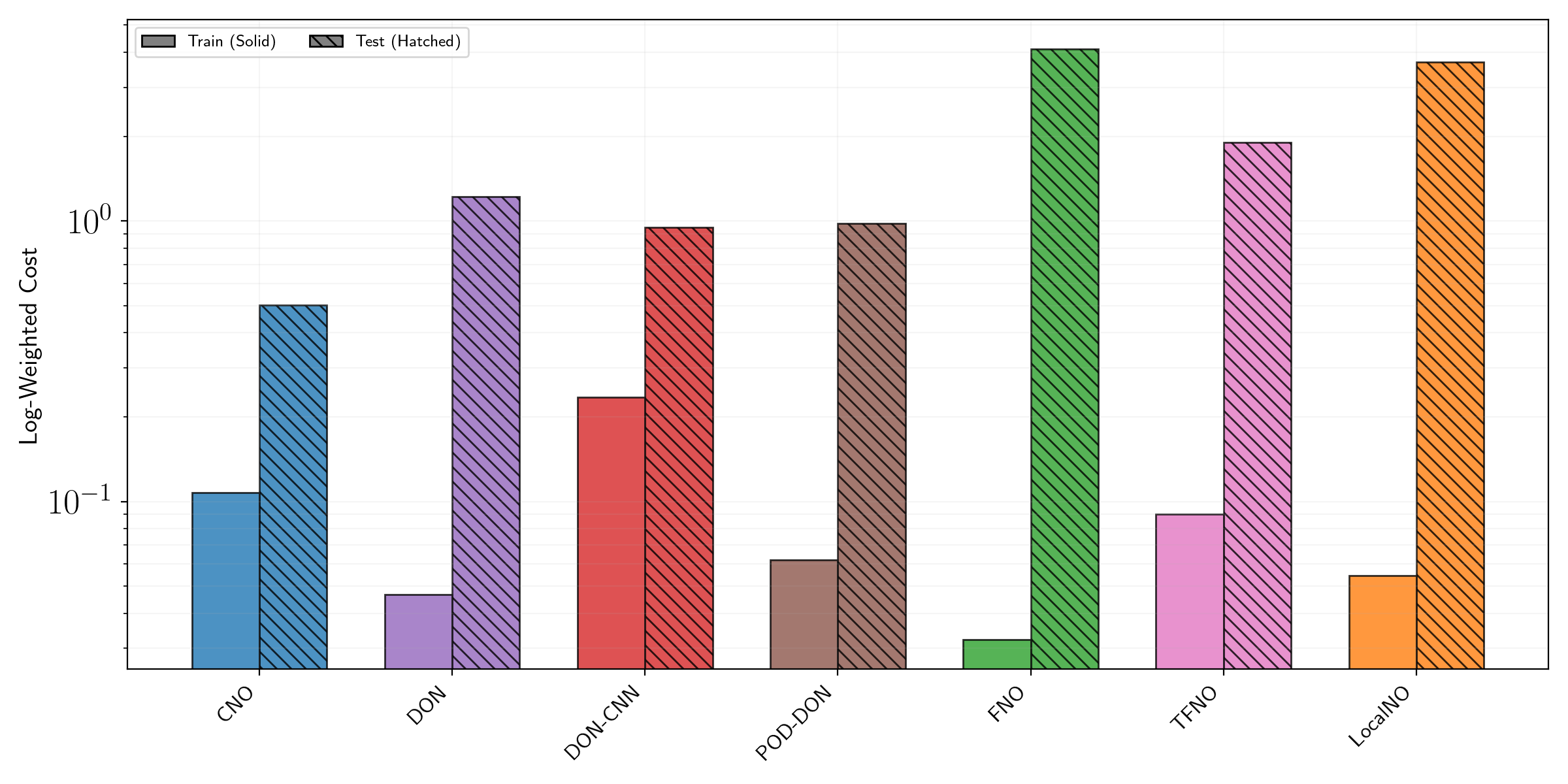}
    \caption{Histogram of the $C_{log}$ Cost for each architecture. The cost related to the training error is shown in solid color, while the test is indicated by the hatched pattern.}
    \label{fig:histo_log}
\end{figure}

As illustrated in Figure \ref{fig:Cost}, the FNO has the highest overall cost, primarily due to its large number of trainable parameters (151 M). On the other hand, DON and their variants have the lowest computational cost because of their limited number of parameters and fast training. However, it is important to note that despite their efficiency, they have difficulty in reproducing the system's translated dynamics. The CNO has high costs in both training and test, largely due to its intensive computational requirements during training (10 h). Due to the significant differences in scale among these models, we adopted the log-weighted metric $\mathcal{C}_{\text{log}}$. This metric mitigates the significant scale disparities among these models. This ensures a more balanced comparison between compact architectures and over-parameterized models. Figure \ref{fig:histo_log} shows the cost function $\mathcal{C}_{log}$ values for training and test errors of each architecture


Figure \ref{fig:histo_log} illustrates the log-weighted cost ($\mathcal{C}_{\text{log}}$), which provides a more fair comparison of the evaluated architectures. The FNO achieves a competitive score because its low training time and training error effectively mitigate the cost associated with its large number of trainable parameters. However, in the test dataset, the FNO and its variants (TFNO and LocalNO) have the highest costs because they cannot maintain reliable performance for the translated dynamics. Similarly, Deep Operator Networks (DONs, DONs-CNN, and POD-DONs) demonstrate consistently high costs for the test dataset, reflecting the difficulty of the generalization for the out-of-distribution scenario. However these models exhibit a lower cost for the training set due to their fast training and limited number of trainable parameters. However, DON-CNN incurs a slightly higher cost than its counterparts due to the added complexity introduced by its encoders. In contrast, the CNO exhibits consistent performance in both training and testing. This stability highlights its unique ability to reliably capture translated dynamics, being the only architecture to do so effectively. Its total cost remains comparable to that of other architectures, mainly due to the high computational cost of training. A final key metric evaluated is the inference latency for a single test instance, compared against the time required to generate a high-fidelity reference solution. This benchmark solution was produced using the Firedrake library \cite{FiredrakeUserManual} on a laptop equipped with an Intel(R) Core(TM) i7-14700HX CPU.

\begin{figure}[!ht]
    \centering
    \begin{subfigure}[t]{0.49\textwidth}
        \centering
        \includegraphics[width=0.9\textwidth]{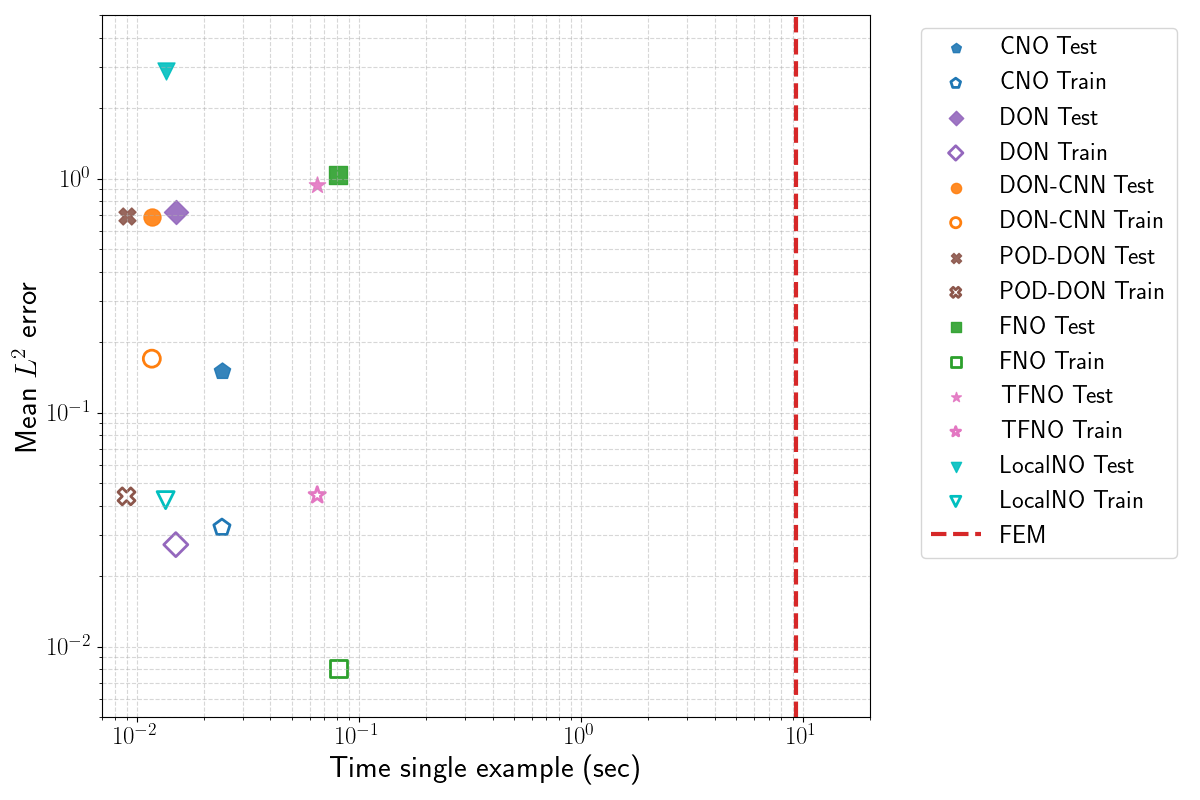}
        \caption{Mean relative $L^2$ error vs. inference time.}
        \label{fig:error_time_mean}
    \end{subfigure}
    \hfill
    \begin{subfigure}[t]{0.49\textwidth}
        \centering
        \includegraphics[width=0.9\textwidth]{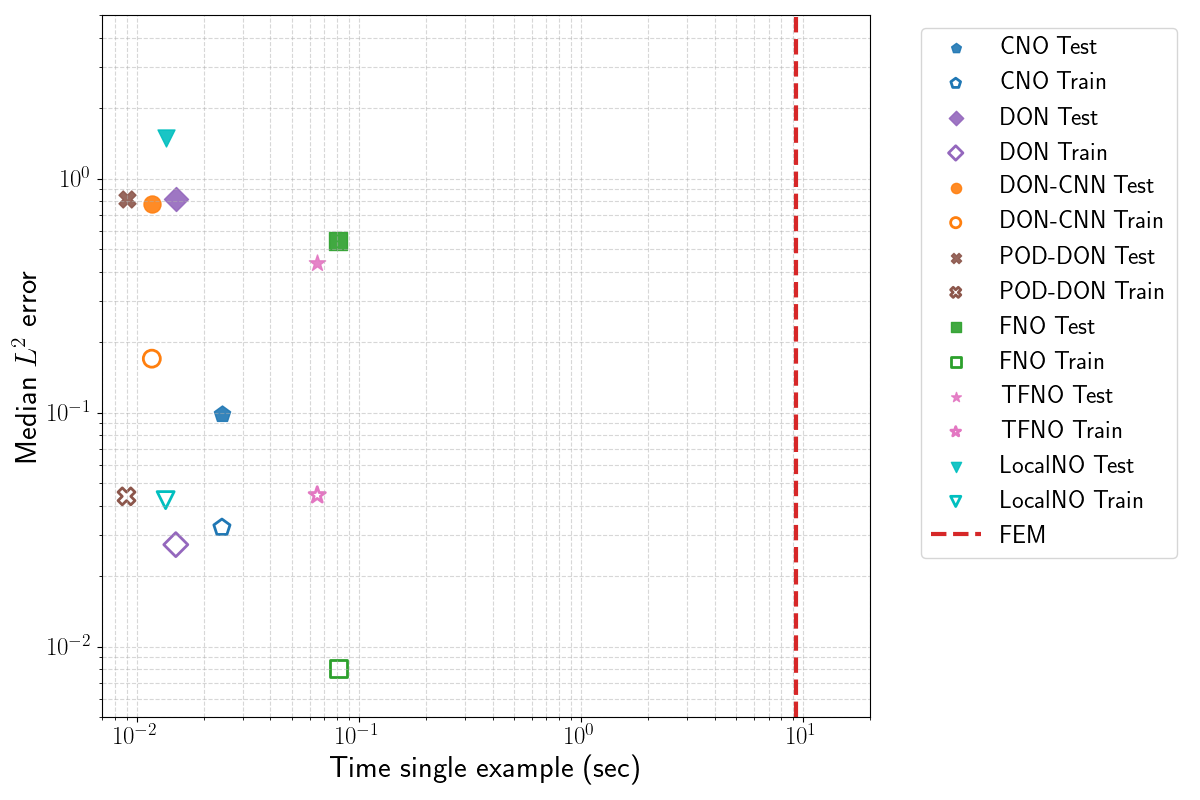}
        \caption{Median relative $L^2$ error vs. inference time.}
        \label{fig:error_time_median}
    \end{subfigure}
    \caption{Comparison of error and inference time for the different architectures. The plots show the inference time relative to the high-fidelity solution time, alongside the accuracy achieved on the training and test sets.}
    \label{fig:error_comparison}
\end{figure}

From Figure \ref{fig:error_comparison}, we can observe that a clear trade-off exists between training and test precision and inference efficiency. While the FNO achieves the lowest training error, it is the most computationally expensive architecture during inference. This limitation is also observed in the TFNO variant. DON and their variants, however, emerge as the most efficient models for real-time evaluation. The CNO and LocalNO offer a balanced middle ground. However, we must note that the LocalNO achieves competitive inference speeds due to its limited number of trainable parameters.
\clearpage
\section{Conclusion}
\label{sec:conclusion}
In this study, we evaluated several Neural Operator architectures, including CNOs, DONs, DONs-CNNs, POD-DONs, FNOs, TFNOs, and LocalNOs, to determine their ability to learn the stiff dynamics and the  translation invariance property of the FitzHugh-Nagumo model. We benchmarked these architectures based on their predictive accuracy in training and test datasets, computational efficiency for the training, and inference speed. Our findings reveal that CNOs are the only models capable of accurately reproducing translated dynamics in the test set, achieving a median $L^2$ relative error of 0.09. However, this robustness comes at the cost of a higher computational effort for the training, requiring approximately 10 hours for a model with 8.2M trainable parameters. In contrast, despite having a significantly larger number of trainable parameters (151.1M), FNOs required a lower training time (4 hours) than CNOs. However, FNOs and their variants have given less accurate predictions for the translated test cases, with many examples yielding high errors. Furthermore, FNOs and TFNOs suffer from substantial inference costs. The LocalNO variant achieved faster inference, though this was primarily due to its significantly lower number of trainable parameters. Regarding DONs and their variants (DONs-CNN and POD-DONs), we observed that they did not generalize well in the translated dynamics of the test set. Nevertheless, DONs architectures demonstrated superior computational efficiency, providing the fastest training and inference speeds among all the models tested. Furthermore, their architecture offers a distinct advantage in terms of flexibility, because they can naturally accommodate non-square geometries without requiring structural modifications. In terms of the training set accuracy, FNOs had the best performance, achieving a relative $L^2$ training error of 0.0081. The accuracy of the other architectures was lower though similar to each other. 
However, our analysis revealed a critical challenge common to all evaluated NOs: determining whether the applied current is sufficient to trigger an action potential, a phenomenon known as the threshold effect. This threshold is difficult to identify because there is no single, precise value for the current. Rather, it depends on the stimulated area and the stimulus duration. For example, incorrectly predicting an action potential when the stimulus is insufficient creates a significant outlier and degrades the NOs ability to accurately capture the correct system dynamics. Future research will focus on improving the ability of NOs to correctly identify whether a stimulus is strong enough to initiate an action potential. A key limitation of this work is that the architectures employed are not intrinsically translation-invariant, nor were specific loss terms introduced. Furthermore, this study does not differentiate between spatial and temporal translation invariance, leaving these as future research topics. In summary, this work provides a rigorous benchmark for NOs, highlighting the trade-offs between computational efficiency and the capacity to learn the stiff dynamics of the FHN model and its translation invariance property.
\clearpage

\section*{Acknowledgements}
   The author would like to thank Massimiliano Ghiotto for helpful discussions regarding Neural Operators and his library, and Luca Franco Pavarino for insightful discussions on ionic models. The author has been supported by grants of MIUR PRIN P2022B38NR, funded by European Union -
Next Generation EU, and by the Istituto Nazionale di Alta Matematica (INdAM - GNCS), Italy. 

\section*{Author Contributions}
    LP: Conceptualization, Methodology, Software, Validation, Formal analysis, Investigation, Data Curation, Writing - Original Draft, Writing - Review \& Editing, Visualization.

\section*{Declarations}
\textbf{Competing Interests} The author declares no competing interests.

\section*{Data Availability}
The code and dataset will be made available on GitHub and Zenodo following the publication of the paper.

\printbibliography

\clearpage
\appendix
\section{Model description and dataset generation}
This section details the mathematical model under consideration, the dataset generation process, and the range of values employed for the applied current $I_{app}$.

\subsection{FitzHugh-Nagumo Model}\label{sec:FHN model}
In this study, we considered the one-dimensional FitzHugh-Nagumo model \cite{franzone2014mathematical,fitzhugh1961impulses}, which is a widely used benchmark for capturing the essential dynamics of excitable cells \cite{cebrian2024six}. The system is governed by the following parabolic reaction-diffusion equations:

\begin{equation}
    \begin{cases}
        \chi C_m \dfrac{\partial V}{\partial t} - \text{div}(D\nabla V) - \chi bV(V- c)(\delta - V) + \beta w = I_{app}, & (x,t)\in [0,1]\times[0,40 ms], \\
        \dfrac{\partial w}{\partial t} = \eta V - \gamma w,                                                              & (x,t)\in [0,1]\times[0,40 ms],
    \end{cases}
    \label{eq:FHN}
\end{equation}

where $V$ represents the transmembrane potential and $w$ is the recovery variable. The parameters used in \eqref{eq:FHN} are summarized in Table \ref{table:fhn parameters}.

\begin{table}[ht!]
    \centering
    \begin{tabular}{lccccccccc}
        \hline
        Parameter & $\chi$ & $C_m$ & $D$   & $b$ & $c$ & $\delta$ & $\beta$ & $\eta$ & $\gamma$ \\
        \hline
        Value     & 1      & 1     & 0.001 & 5   & 0.1 & 1        & 1       & 0.1    & 0.25     \\
        \hline
    \end{tabular}
    \caption{Parameters assigned to the FHN model in \eqref{eq:FHN}.}
    \label{table:fhn parameters}
\end{table}
\subsection{Dataset generation}\label{app_sec:dataset}
The dataset was generated using Firedrake \cite{FiredrakeUserManual}, a finite element library. For the spatial discretization, second-degree Lagrange polynomials were employed over a mesh consisting of 200 intervals. Temporal integration was performed using the forward Euler method with a constant time step of $\Delta t = 0.05$. We solved the resulting nonlinear systems using the Newton method, and we handled the underlying linear systems using the generalized minimal residual method. To ensure efficient convergence, we utilized a generalized additive Schwarz method as a preconditioner. For the applied current defined in \eqref{eq:applied_current},  the stimulated region is defined as:
\[\Omega_{T,stim} = [x_{\min}, x_{\min} + 0.04] \times [T_{\min}, T_{\min} + 1].\]
The parameter ranges and values used for the training, validation, and test datasets are summarized in Table \ref{table:fhn_dataset}.

\begin{table}[!ht]
    \centering
    \resizebox{\textwidth}{!}{
        \begin{tabular}[t]{cccccc}
            \hline
            Name              & \multicolumn{2}{c}{Range of values of $i$} & Time of the stimulus & Position of the stimulus & Number of examples        \\
                              & min                                        & max                  & $T_{min}$                & $x_{min}$          &      \\
            \hline
            Train             & 0.1                                        & 3                    & 5                        & 0-0.96             & 2000 \\
            Validation / Test & 0.1                                        & 3                    & 0-37                     & 0-0.96             & 500  \\
            \hline
        \end{tabular}}
    \caption{FHN model: Range of values of the training, validation and test set. }
    \label{table:fhn_dataset}
\end{table}

\section{Automatic hyperparameter tuning}\label{app_sec:ray}
In this section, we provide details on the automatic hyperparameter tuning used for each of the architectures considered in this work. We performed 100 distributed and parallel trials for each architecture,  employing an automatic hyperparameter optimization algorithm implemented in the HyperOptSearch library. In particular, we employed the Tree-structured Parzen Estimator algorithm \cite{hyperopt12bergstra}, and the ASHA scheduler \cite{AHSHA}  to automatically stop trials that were not satisfactory. Each model was trained using the data-driven relative $L^2$ norm as the loss function.
\[
    \begin{split}
        \text{Loss}\left( u,\, \mathcal{G}_\theta(I_{app})\right) & = \frac{1}{2} \sum_{j=1}^{2}\frac{\|u_j - \mathcal{G}_\theta(I_{app})_j\|_{L^2(\Omega_T)}}{\|u_j\|_{L^2(\Omega_T)}}                                                                                          \\
                                                                  & \approx \frac{1}{2} \sum_{j=1}^{2} \frac{\left(\sum_{k=1}^{n_{points}} |u_j(x_k,t_k) - \mathcal{G}_\theta(I_{app})_j(x_k,t_k)|^2\right)^{1/2}}{\left(\sum_{k=1}^{n_{points}} |u_j(x_k,t_k)|^2\right)^{1/2}},
    \end{split}
\]
where $u$ is the solution of the system \eqref{eq:monodomain} and $\mathcal{G}_\theta$ is the operator defined in \eqref{eq:neural_operator_map}. This leads to the following optimization problem:
\[
    \underset{\theta\in\Theta}{\arg\min} \ \ \frac{1}{n_{train}} \sum_{i=1}^{n_{train}} \text{Loss}\left( u^i,\, \mathcal{G}_\theta(I_{app}^i)\right),
\]
where $\{(I_{app}^i, u^i)\}_{i=1}^{n_{train}}$ is the dataset consisting of input-output pairs.

\subsection{Convolutional Neural Operators hyperparameters}\label{app_subsec:CNO}
Table \ref{table:CNO hyperparams} summarizes the hyperparameters for the CNOs \cite{CNO23raonic}, detailing the configuration of the search space and the optimal values identified by the automatic hyperparameter tuning.
\begin{table}[h!]
    \centering
    \resizebox{\textwidth}{!}{
        \begin{tabular}{llll}
            \hline
            \textbf{Hyperparameter} & \textbf{Description}                    & \textbf{Search Space}               & \textbf{Found} \\
            \hline
            \multicolumn{4}{c}{\textbf{Architecture}}                                                                                \\
            $N_{\text{layers}}$     & Number of layers of the U-Net           & $1\leq N_{\text{layers}}\leq5$      & 4              \\
            C                       & Dimension of channel multiplier         & C$\in \{8, 16, 24, 32, 40, 48\}$    & 32             \\
            $N_{\text{res,neck}}$   & Number residual block of the last layer & $1\leq N_{\text{res,neck}} \leq 6$  & 2              \\
            $N_{\text{res}}$        & Number residual block                   & $1\leq N_{\text{res}}\leq 8$        & 7              \\
            $k$                     & Kernel size                             & $k \in \{3,5,7\}$                   & 3              \\
            \hline
            \multicolumn{4}{c}{\textbf{Optimizer}}                                                                                   \\
            \hline
            $\learningRate$         & Initial learning rate                   & 1e-4 $\leq \learningRate\leq $1e-2  & 1.9e-4         \\
            $\weightDecay$          & Weight decay regularization factor      & 1e-6 $\leq \weightDecay\leq$1e-3    & 3.2e-3         \\
            $\schedulerGamma$       & Rate scheduler factor                   & $0.75\leq \schedulerGamma\leq 0.99$ & 0.91           \\
            \hline
        \end{tabular}}
    \caption{CNOs hyperparameters and the search space used for hyperparameter tuning.}
    \label{table:CNO hyperparams}
\end{table}

\subsection{Deep Operator Network hyperparameters}\label{app_subsec:DON}
The DONs implementation is taken from the DeepXDE library \cite{lu2021deepxde}. Table \ref{table:DON_hyperparams} summarizes the hyperparameters for the DONs \cite{DON21lu}, detailing the configuration of the search space and the optimal values identified by the automatic hyperparameter tuning.

\begin{table}[h!]
    \resizebox{\textwidth}{!}{
        \begin{tabular}{llll}
            \hline
            \textbf{Hyperparameter} & \textbf{Description}      & \textbf{Search Space}                      & \textbf{Found} \\
            \hline
            $p$                     & Number of basis functions & $200 \leq p \leq 800$                      & 762            \\
            \hline
            \multicolumn{4}{c}{\textbf{Branch Network}}                                                                       \\
            \hline
            $N_{branch}$            & Number of hidden layers   & $1 \leq N_{branch} \leq 6$                 & 6              \\
            $W_{branch}$            & Network width (neurons)   & $64 \leq W_{branch} \leq 200$              & 124            \\
            $\sigma_{branch}$       & Activation function       & $\sigma_{branch} \in$ \{relu, gelu, silu\} & relu           \\
            \hline
            \multicolumn{4}{c}{\textbf{Trunk Network}}                                                                        \\
            \hline
            $N_{trunk}$             & Number of hidden layers   & $1 \leq N_{trunk} \leq 6$                  & 4              \\
            $W_{trunk}$             & Network width (neurons)   & $64 \leq W_{trunk} \leq 200$               & 148            \\
            $\sigma_{trunk}$        & Activation function       & $\sigma_{trunk} \in$ \{relu, gelu, silu\}  & relu           \\
            \hline
            \multicolumn{4}{c}{\textbf{Optimizer}}                                                                            \\
            \hline
            $\learningRate$         & Initial learning rate     & $10^{-4} \leq \learningRate \leq 10^{-2}$  & 1.8e-3         \\
            $\weightDecay$          & Weight decay factor       & $10^{-6} \leq \weightDecay \leq 10^{-3}$   & 4.5e-4         \\
            $\schedulerGamma$       & Rate scheduler factor     & $0.75 \leq \schedulerGamma \leq 0.99$      & 97             \\
            \hline
        \end{tabular}}
    \caption{DONs hyperparameters and the search space used for  hyperparameter tuning.}
    \label{table:DON_hyperparams}
\end{table}
\subsection{Deep Operator Network with CNN encoders hyperparameters}\label{app_subsec:DON_CNN}
Table \ref{table:DON_CNN_hyperparams} summarizes the hyperparameters for the DONs-CNN, detailing the configuration of the search space and the optimal values identified by the automatic hyperparameter tuning.

\begin{table}[h!]
    \centering
    \resizebox{\textwidth}{!}{
        \begin{tabular}{llll}
            \hline
            \textbf{Hyperparameter} & \textbf{Description}                      & \textbf{Search Space}                                                        & \textbf{Found} \\
            \hline
            $p$                     & Number of basis functions                 & $200\leq p \leq 500$                                                         & 332            \\
            \hline
            \multicolumn{4}{c}{\textbf{Branch Network}}                                                                                                                         \\
            \hline
            $N_{conv}$              & Number of convolutional layers            & $3\leq N_{conv}\leq7$                                                        & 3              \\
            $C_{start}$             & Starting number of channels (first layer) & $C_{start} \in \{10, 20, 30, 40\}$                                           & 40             \\
            $C_{step}$              & Channel increment per layer               & $C_{step} \in \{10, 20, 30\}$                                                & 30             \\
            $C_{ch}$                & Channel configuration                     & -                                                                            & [40, 70, 100]  \\
            $N_{FNN, branch}$       & Number of layers for the FNN              & $2\leq N_{FNN, branch}\leq 7$                                                & 3              \\
            $N_{neuron, branch}$    & Number of neurons per dense layer         & $N_{neuron, branch} \in \{16, 32, 64, 128, 256\}$                            & 16             \\
            $\sigma_{branch}$       & Activation function                       & $\sigma_{branch}\in \{\text{tanh, relu, leaky\_relu, sigmoid, silu, gelu}\}$ & relu           \\
            Normalization           & Type of layer normalization               & Normalization $\in \{\text{none, batch, layer}\}$                            & none           \\
            \hline
            \multicolumn{4}{c}{\textbf{Trunk Network}}                                                                                                                          \\
            \hline
            $N_{FNN, trunk}$        & Number of layers for the FNN              & $2\leq N_{FNN, trunk} \leq 7$                                                & 4              \\
            $N_{neuron, trunk}$     & Number of neurons per dense layer         & $N_{neuron, trunk}\in \{16, 32, 64, 128, 256\}$                              & 16             \\
            $\sigma_{trunk}$        & Activation function                       & $\sigma_{trunk}\in \{\text{tanh, relu, leaky\_relu, sigmoid}\}$              & leaky\_relu    \\

            Architecture            & Use of residual connections               & Classic or Residual                                                          & Classical      \\
            Layer Norm              & Apply layer normalization                 & True or False                                                                & False          \\
            \hline
            \multicolumn{4}{c}{\textbf{Optimizer}}                                                                                                                              \\
            \hline
            $\learningRate$         & Initial  learning rate                    & 1e-4$\leq \learningRate \leq$ 1e-2                                           & 6.6e-3         \\
            $\weightDecay$          & Weight decay regularization factor        & 1e-6$\leq \weightDecay \leq $ 1e-3                                           & 4.6e-4         \\
            $\schedulerGamma$       & Rate scheduler factor                     & $0.75 \leq \schedulerGamma \leq 0.99$                                        & 0.84           \\
            \hline
        \end{tabular}}
    \caption{DONs-CNN hyperparameters and the search space used for hyperparameter tuning}
    \label{table:DON_CNN_hyperparams}
\end{table}

\subsection{Proper Orthogonal Decomposition DeepONets hyperparameters}\label{app_subsec:POD_DON}
The POD-DONs implementation is taken from the DeepXDE library \cite{lu2021deepxde}. Table \ref{table:POD_DON_hyperparams} summarizes the hyperparameters for the POD-DONs \cite{lu2022comprehensive}, detailing the configuration of the search space and the optimal values identified by the automatic hyperparameter tuning.
\begin{table}[h!]
    \centering
    \resizebox{\textwidth}{!}{
        \begin{tabular}{llll}
            \hline
            \textbf{Hyperparameter} & \textbf{Description}                   & \textbf{Search Space}                      & \textbf{Found}        \\
            \hline
            $p$                     & Number of basis functions ($n\_basis$) & $200 \leq p \leq 800$                      & 500                   \\
            \hline
            \multicolumn{4}{c}{\textbf{Branch Network}}                                                                                           \\
            \hline
            $N_{branch}$            & Number of hidden layers                & $1 \leq N_{branch} \leq 6$                 & 6                     \\
            $W_{branch}$            & Network width                          & $64 \leq W_{branch} \leq 200$              & 92                    \\
            $\sigma_{branch}$       & Activation function                    & $\sigma_{branch} \in$ \{relu, gelu, silu\} & gelu                  \\
            \hline
            \multicolumn{4}{c}{\textbf{Optimizer}}                                                                                                \\
            \hline
            $\learningRate$         & Initial learning rate                  & 1e-4$ \leq \learningRate \leq $1e-2        & $6.7 \times 10^{-4}$  \\
            $\weightDecay$          & Weight decay factor                    & 1e-6$ \leq \weightDecay \leq $1e-3         & $4.25 \times 10^{-4}$ \\
            $\schedulerGamma$       & Rate scheduler factor                  & $0.75 \leq \schedulerGamma \leq 0.99$      & 0.92                  \\
            \hline
        \end{tabular}}
    \caption{POD-DONs hyperparameters and the search space used for hyperparameter tuning.}
    \label{table:POD_DON_hyperparams}
\end{table}

\subsection{Fourier Neural Operators hyperparameters}\label{app_subsec:FNO}
Table \ref{table:FNO_hyperparams} summarizes the hyperparameters for the FNOs \cite{FNO20li}, detailing the configuration of the search space and the optimal values identified by the automatic hyperparameter tuning.
\begin{table}[h!]
    \centering
    \resizebox{\textwidth}{!}{
        \begin{tabular}{llll}
            \hline
            \textbf{Hyperparameter} & \textbf{Description}                                                & \textbf{Search Space}                                   & \textbf{Found} \\
            \hline
            \multicolumn{4}{c}{\textbf{Architecture}}                                                                    \\
            \hline
            $\width$                & Hidden dimension                                                    & $\width \in \{4,8,16,32,64,96,128,160,192,220\}$        & 128            \\
            $\hiddenLayer$          & Number of hidden layers                                             & 1 $\leq\hiddenLayer \leq$6                              & 4              \\
            $\fourierModes$         & Number of Fourier modes                                             & $\fourierModes \in \{2, 4, 8, 12, 16, 20, 24, 28, 32\}$ & 24             \\
            $\activation$           & Activation function                                                 & $\activation\in$\{tanh, relu, gelu, leaky\_relu\}       & relu           \\
            $\paddingPoints$        & Number of padding points per direction                              & 1$\leq \paddingPoints \leq$16                           & 1              \\
            Architecture            & Fourier architecture \eqref{eq:classic}, \eqref{eq:mlp} or residual & Classic, MLP or Residual                                & Classic        \\
            \hline
            \multicolumn{4}{c}{\textbf{Optimizer}}                                                                                                                                   \\
            \hline
            $\learningRate$         & Initial learning rate                                               & 1e-4 $\leq\learningRate \leq$ 1e-2                      & 1.8e-3         \\
            $\weightDecay$          & Weight decay regularization factor                                  & 1e-6 $\leq\weightDecay\leq$ 1e-3                        & 4e-6           \\
            $\schedulerGamma$       & Rate scheduler factor                                               & $0.75\leq \schedulerGamma \leq 0.99$                    & 0.85           \\
            \hline
        \end{tabular}}
    \caption{FNOs hyperparameters and the search space used for hyperparameter tuning.}
    \label{table:FNO_hyperparams}
\end{table}

\subsection{Tucker Tensorized FNOs hyperparameters}\label{app_subsec:TFNO}
The TFNOs implementation is taken from the Neural Operator library \cite{kossaifi2025librarylearningneuraloperators}. Table \ref{table:TFNO_hyperparams} summarizes the hyperparameters for the TFNOs \cite{zhou2026tuckerfno}, detailing the configuration of the search space and the optimal values identified by the automatic hyperparameter tuning.

\begin{table}[h!]
    \centering
    \resizebox{\textwidth}{!}{
        \begin{tabular}{llll}
            \hline
            \textbf{Hyperparameter} & \textbf{Description}    & \textbf{Search Space}                         & \textbf{Found} \\
            \hline
            \multicolumn{4}{c}{\textbf{Architecture}}                                                                          \\
            \hline
            $\width$                & Hidden dimension        & $\width \in \{4, 8, 16, \dots, 192\}$         & 192            \\
            $\hiddenLayer$          & Number of hidden layers & $1 \leq \hiddenLayer \leq 5$                  & 1              \\
            $\fourierModes$         & Number of Fourier modes & $\fourierModes \in \{2, 4, 8, \dots, 32\}$    & 8              \\
            $\activation$           & Activation function     & $\activation \in$ \{gelu, relu, leaky\_relu\} & leaky\_relu    \\
            Domain Padding          & Domain padding fraction & $\{0, 0.1, 0.2\}$                             & 0.2            \\
            MLP skip                & Skip connection in MLP  & \{Linear, identity, soft-gating, None\}       & identity       \\
            FNO skip                & Skip connection in FNO  & \{Linear, identity, soft-gating, None\}       & soft-gating    \\
            MLP                     & Usage of Channel MLP    & True or False                                 & False          \\
            Rank                    & Tensor rank ratio       & $\{0.05, 0.1, 0.2, 0.5\}$                     & 0.2            \\

            \hline
            \multicolumn{4}{c}{\textbf{Optimizer}}                                                                             \\
            \hline
            $\learningRate$         & Initial learning rate   & 1e-4$ \leq \learningRate \leq $1e-2           & 5.5e-3         \\
            $\weightDecay$          & Weight decay factor     & 1e-6$ \leq \weightDecay \leq $ 1e-3           & 5.5e-4         \\
            $\schedulerGamma$       & Rate scheduler factor   & $0.75 \leq \schedulerGamma \leq 0.99$         & 0.8            \\
            \hline
        \end{tabular}}
    \caption{TFNOs hyperparameters and the search space used for hyperparameter tuning.}
    \label{table:TFNO_hyperparams}
\end{table}

\subsection{Local Neural Operators hyperparameters}\label{app_subsec:LocalNO}
The LocalNOs implementation is taken from the Neural Operator library \cite{kossaifi2025librarylearningneuraloperators}. Table \ref{table:LocalNO_hyperparams} summarizes the hyperparameters for the LocalNOs \cite{liu2024neural}, detailing the configuration of the search space and the optimal values identified by the automatic hyperparameter tuning.
\begin{table}[h!]
    \centering
    \resizebox{\textwidth}{!}{
        \begin{tabular}{llll}
            \hline
            \textbf{Hyperparameter}  & \textbf{Description}               & \textbf{Search Space}                                & \textbf{Found} \\
            \hline
            \multicolumn{4}{c}{\textbf{Architecture}}                                                                                             \\
            \hline
            $\width$                 & Hidden dimension                   & $\width \in \{4,8,16,32,64,96,128,160,192\}$         & 8              \\
            $\hiddenLayer$           & Number of hidden layers            & 1 $\leq\hiddenLayer \leq$6                           & 2              \\
            $\fourierModes$          & Number of Fourier modes            & $\fourierModes \in \{2,4,8,12, 16, 20, 24, 28, 32\}$ & 24             \\
            $\activation$            & Activation function                & $\activation\in$\{relu, gelu, leaky\_relu\}          & relu           \\
            Padding                  & Type of  padding                   & reflect or zeros                                     & zeros          \\
            Factorization            & Type of Factorization              & None, Tucker, CP, TT                                 & TT             \\
            MLP skip                 & Skip connection in MLP             & Linear, identity and soft-gating                     & linear         \\
            local skip               & Skip connection in LocalNO         & Linear, identity and soft-gating                     &  identity      \\
            Differential kernel size & Finite difference kernel size      & \{3,5,7\}                                            & 7              \\
            MLP                      & Usage of MLP                       & True or False                                        & True           \\
            Rank                     & Tensor rank for factorization      & \{0.05,0.1,0.2,0.5,1\}                               & 0.1            \\
            Disco Layer              & Usage of Disco (integral) layer    & True or False                                        & False          \\

            \hline
            \multicolumn{4}{c}{\textbf{Optimizer}}                                                                                                \\
            \hline
            $\learningRate$          & Initial learning rate              & 1e-4 $\leq\learningRate \leq$ 1e-2                   & 1.4e-3         \\
            $\weightDecay$           & Weight decay regularization factor & 1e-6 $\leq\weightDecay\leq$ 1e-3                     & 2.8e-4         \\
            $\schedulerGamma$        & Rate scheduler factor              & $0.75\leq \schedulerGamma \leq 0.99$                 & 0.97           \\
            \hline
        \end{tabular}}
    \caption{LocalNOs hyperparameters and the search space used for hyperparameter tuning.}
    \label{table:LocalNO_hyperparams}
\end{table}



\end{document}